\documentclass[journal,twoside,web]{ieeecolor}

\usepackage{wrapfig}
\usepackage{soul}
\usepackage{rotating}
\usepackage{chngcntr}
\usepackage{apptools}
\usepackage{listings}
\usepackage{placeins}
\AtAppendix{\counterwithin{thm}{section}}
\usepackage{graphicx}
\usepackage{graphics}
\usepackage{amssymb}

\usepackage{amsmath}
\usepackage[colorlinks=true,linkcolor=.]{hyperref}
\usepackage{xspace}
\usepackage{algpseudocode}
\usepackage{bbm}
\usepackage{algorithmicx}
\usepackage{subfig}
\usepackage{float}
\usepackage{cite}
\usepackage[normalem]{ulem}
\usepackage{tikz,pgfplots}
\usepackage{siunitx}
\usetikzlibrary{shapes,arrows}
\usetikzlibrary{decorations.pathreplacing,decorations.markings}
\usetikzlibrary{positioning,calc}
\pgfplotsset{compat=1.7}

\tikzstyle{block}=[draw opacity=0.7,line width=1.4cm]

\tikzset{block/.style={%
        inner xsep=1mm,
        inner ysep=1.5mm,
        rectangle,very thick,draw}}
\tikzset{sum/.style={%
        circle,
        minimum size=2mm,inner xsep=1.2mm,inner ysep=1.2mm,
        very thick,draw}}
\tikzset{point/.style={%
        minimum size=0mm,inner xsep=4mm,inner ysep=0mm,draw}}
\tikzset{link/.style={->,very thick,>=stealth}}
\tikzset{undirlink/.style={<->,very thick,>=stealth}}
\tikzset{pole/.style={cross out, draw=black, minimum size=2*(#1-\pgflinewidth), inner sep=0pt, outer sep=0pt},cross/.default={1pt}}

\usepackage{xcolor}
\usepackage{pifont}
\usepackage{rotating}
\usepackage{sidecap}
\usepackage{booktabs}
\usepackage{comment}
\usepackage{algorithm}
\usepackage{xcolor}
\usepackage{pifont}

\newcommand{\cmark}{\textcolor{green!60!black}{\checkmark}}
\newcommand{\xmark}{\textcolor{red!70!black}{\texttimes}}

\makeatletter
\algrenewcommand\alglinenumber[1]{\scriptsize #1:} 
\makeatother

\title{\LARGE \bf
FedMPDD: Communication-Efficient Federated~Learning with Privacy Preservation Attributes via Projected Directional Derivative}

\author{Mohammadreza Rostami and Solmaz S. Kia, \emph{Senior Member, IEEE}
  \thanks{The authors are with the Department of Mechanical and Aerospace Engineering, University of California Irvine, Irvine, CA 92697,  
    {\tt\small \{mrostam2,solmaz\}@uci.edu}. } %
}

\definecolor{subsectioncolor}{RGB}{0,70,140}

\newcommand{\reza}[1]{{\color{blue}#1}}

\newcommand{\vect}[1]{\boldsymbol{\mathbf{#1}}}

 \newcommand{\boxend}{\hfill \ensuremath{\Box}}

\newtheorem{thm}{Theorem}

\newtheorem{lem}{Lemma}

\newtheorem{assump}{Assumption}

\newtheorem{definition}{Definition}
\newtheorem{remark}{Remark}
\newcommand{\oprocendsymbol}{\hbox{$\bullet$}}
\newcommand{\oprocend}{\relax\ifmmode\else\unskip\hfill\fi\oprocendsymbol}

\makeatletter
\renewcommand*{\@opargbegintheorem}[3]{\trivlist
      \item[\hskip \labelsep{ #1\ #2}] (#3):\ \itshape}
\makeatother

\usepackage{setspace}

\begin{document}
\maketitle

\thispagestyle{empty}
\pagestyle{empty}
\begin{abstract}
This paper introduces \texttt{FedMPDD} (\textbf{Fed}erated Learning via \textbf{M}ulti-\textbf{P}rojected \textbf{D}irectional \textbf{D}erivatives), a novel algorithm that simultaneously optimizes bandwidth utilization and enhances privacy in Federated Learning. 
The core idea of \texttt{FedMPDD} is to encode each client's high-dimensional gradient by computing its directional derivatives along multiple random vectors. This compresses the gradient into a much smaller message, significantly reducing uplink communication costs from $\mathcal{O}(d)$ to $\mathcal{O}(m)$, where $m \ll d$. The server then decodes the aggregated information by projecting it back onto the same random vectors. Our key insight is that averaging multiple projections overcomes the dimension-dependent convergence limitations of a single projection. We provide a rigorous theoretical analysis, establishing that \texttt{FedMPDD} converges at a rate of $\mathcal{O}(1/\sqrt{K})$, 
matching the performance of FedSGD. Furthermore, we demonstrate that our method provides some inherent privacy against gradient inversion attacks due to the geometric properties of low-rank projections, offering a tunable privacy-utility trade-off controlled by the number of projections. Extensive experiments on benchmark datasets validate our theory and demonstrates our results. 
\end{abstract}
\section{Introduction}\label{sec::Intro}
Federated Learning (FL) is a foundational paradigm for collaboratively training models across $N$ edge devices by leveraging their local computational resources~\cite{mcmahan2017communication, kairouz2021advances, chen2019deep}. Beyond traditional machine learning applications, FL has been extended to control systems, including model-free learning with heterogeneous dynamical systems~\cite{zeng2021federated} and federated linear quadratic regulator  approaches~\cite{wang2023model}. FL seeks to solve the distributed optimization problem
\begin{equation}\label{eq::opt_prob}
\text{minimize}_{\vect{x}\in\mathbb{R}^d}~~ f(\vect{x}) = \frac{1}{N} \sum\nolimits_{i=1}^N f_i(\vect{x}),
\end{equation}
where $f_i:\mathbb{R}^d\to\mathbb{R}$ is the local objective (loss) function of client $i$, and $f:\mathbb{R}^d\to\mathbb{R}$ is the global objective. Lacking central access to local objectives, FL iteratively communicates between a server and a client subset $\mathcal{A}_k$ ( $|\mathcal{A}_k|=\beta N$, where \( \beta \in (0, 1] \) denotes the client participation rate). In each round $k$, the server sends out the global model $\vect{x}_k$. Selected clients compute local updates (e.g., mini-batch gradients $\vect{g}_i(\vect{x}_k)$) and transmit them to the server. The server aggregates these gradients by averaging: $\vect{g}(\vect{x}_k) = \frac{1}{\beta N} \sum_{i \in \mathcal{A}_k} \vect{g}_i(\vect{x}_k)$, and updates the global model: $\vect{x}_{k+1} = \vect{x}_k - \eta ~\vect{g}(\vect{x}_k)$, as in \texttt{FedSGD}~\cite{mcmahan2017communication}, where $\eta$ is a suitable learning rate.

A key bottleneck in FL algorithms like \texttt{FedSGD} is the substantial uplink communication overhead from transmitting $d$-dimensional gradients $\vect{g}_i(\vect{x}_k)$ from clients to the server, requiring \(32d\) bits per client per round, assuming single-precision floating-point representation (32 bits per value). For instance, a ResNet-18 model ($\sim 11 \times 10^6$ parameters) necessitates approximately $42$MB transmission per client per round. This high cost severely impacts efficiency in bandwidth-constrained real-world deployments~\cite{chen2021communication, niknam2020federated, shahid2021communication, li2020federated, rostami2024first}. Strategies to reduce FL communication volume or frequency fall into three main classes: \emph{model compression}, \emph{local computation with client selection}, and \emph{gradient compression}. Model compression reduces global model size, for example, by using smaller models with local representations~\cite{liang2020think}. Local computation and client selection decrease communication frequency or the number of clients per round through techniques like multiple local updates~\cite{stich2018local, mcmahan2017communication, karimireddy2020scaffold} and client subset selection~\cite{sattler2019robust, liang2020think}. Gradient compression reduces the size of transmitted gradients using methods like quantization~\cite{alistarh2017qsgd, horvoth2022natural, karimireddy2019error,shlezinger2020uveqfed, bernstein2018signsgd, reisizadeh2020fedpaq, suresh2017distributed}, sparsification~\cite{ivkin2019communication, lin2017deep}, and structured/sketched updates. Structured updates, such as low-rank adaptation~\cite{hu2022lora, yi2023pfedlora, qi2024fdlora, zhang2018systematic, ullrich2017soft, bertsimas2023new, cho2024heterogeneous}, use a pre-defined subspace for parameters, while sketched updates compress gradients using a shared random matrix fixed at initialization via techniques like projection matrices~\cite{park2023regulated, azam2021recycling, guo2024low} or Count-Sketch~\cite{ivkin2019communication, rothchild2020fetchsgd, haddadpour2020fedsketch, jiang2018sketchml}. These approaches rely on a \textit{static} projection for all clients and rounds. While these existing methods provide varying degrees of communication reduction, they often face trade-offs between compression ratio, computational overhead, convergence guarantees, and adaptability to different communication budgets, motivating the need for fundamentally new paradigms that can achieve more aggressive communication reduction without sacrificing convergence performance.

This paper proposes \texttt{FedMPDD} (\underline{Fed}erated learning with \underline{M}ulti-\underline{P}rojected \underline{D}irectional \underline{D}erivatives), a novel framework that fundamentally tackles the critical communication efficiency bottleneck to enable more practical and scalable deployment of FL in bandwidth-constrained environments. Unlike existing gradient compression techniques, our approach introduces a fundamentally new multiplicative encoding paradigm through the \emph{projected directional derivative}---a concept recently explored in optimization contexts including balancing computational cost and memory in deep learning gradient calculations~\cite{fournier2023can, ren2022scaling, silver2021learning, baydin2022gradients, rostami2024projected} and zeroth-order optimization~\cite{nesterov2017random}. The method we propose follows the core structure of \texttt{FedSGD} but achieves dramatic communication reduction by employing the \emph{projected directional derivative} $\hat{\vect{g}}_i(\vect{x}_k)=\vect{u}_{k,i}^\top\vect{g}_i(\vect{x}_k)\vect{u}_{k,i}$ (defined in~\eqref{def:proj}) instead of the full stochastic gradient $\vect{g}_i(\vect{x}_k)$. We decompose this projected directional gradient into a scalar directional derivative 
$\vect{u}_{k,i}^\top \vect{g}_i(\vect{x}_k)$ computed locally by client $i$. Each client independently samples $\vect{u}_{k,i}$, then transmits only two scalars: the directional derivative $\vect{u}_{k,i}^\top \vect{g}_i(\vect{x}_k)$ and the random seed $r_{k,i}$ used to generate $\vect{u}_{k,i}$. On the server side, the received seed $r_{k,i}$ is used to reconstruct the identical $d$-dimensional vector $\vect{u}_{k,i}$, enabling the server to form the gradient estimator $\hat{\vect{g}}_{i}(\vect{x}_{k})$ without ever transmitting the full vector. However, our analysis shows that the noisy gradient estimator from a single projected directional derivative would severely degrade performance. To address this, we propose our main algorithm \texttt{FedMPDD}, which introduces a principled multi-projection aggregation mechanism that averages $m$ projected directional derivatives. Crucially, unlike structured and sketched updates that rely on static projections, our approach uses a \textit{dynamic} projection strategy where the projection directions $\vect{u}_{k,i}$ are randomly and independently sampled for each client $i$ at every round $k$. We show that when $m$ independent projections are aggregated, the classical Johnson--Lindenstrauss guarantees (see details in Section~\ref{sec::Multi-projection}) ensure that the gradient norm is preserved up to small distortion with high probability. As a result, \texttt{FedMPDD} achieves accurate and stable gradient approximation by transmitting only $m+1$ scalars per client per round reducing communication from $32d$ to $32(m+1)$ bits and achieving orders of magnitude savings (e.g., from $42$MB to kilobytes for ResNet-18) while maintaining $\mathcal{O}(1/\sqrt{K})$ convergence rates comparable to full-gradient methods.

While our primary focus is communication efficiency, an interesting emergent property of our encoding scheme is that it naturally provides some level of privacy protection. In FL implementations, Gradient Inversion Attacks (GIAs)~\cite{zhu2019deep, zhao2020idlg, huang2021evaluating, yin2021see, melis2019exploiting, li2022e2egi} (detailed in Appendix~\ref{DLG:explation}) remain a critical security concern, where a curious server can reconstruct raw data from transmitted gradients. Conventional defenses, such as Local Differential Privacy (LDP)~\cite{wei2020federated, truex2020ldp, zhao2020local, seif2020wireless}, rely on injecting noise, which introduces a known trade-off between privacy guarantee and model convergence~\cite{jere2020taxonomy}. Other schemes have combined differential privacy with gradient compression~\cite{amiri2021compressive, lyu2021dp, lang2023joint, agarwal2018cpsgd}, typically through adding Gaussian noise before quantization or integrating DP into quantization schemes, though these approaches often assume a trusted server and introduce convergence-degrading noise. In contrast, the privacy preservation in \texttt{FedMPDD} arises inherently from the geometric properties of our communication reduction mechanism. Specifically, the encoding matrix $\vect{u}_{k, i}\vect{u}_{k, i}^\top$ is rank-deficient ($rank=1 \ll d$), which creates an underdetermined system when attempting to invert the gradient. This nullspace property prevents unique gradient recovery, offering a privacy defense against GIAs that is fundamentally different from additive noise methods and is achieved without sacrificing the communication benefits.

The key contributions of this work are summarized as follows:
\begin{itemize}[leftmargin=10pt]
    \item Significant Communication Reduction: \texttt{FedMPDD} dramatically reduces uplink communication from $\mathcal{O}(d)$ to just $m + 1$ scalars per client per round ($\mathcal{O}(m)$ bits, where $m \ll d$). 

    \item Fast Convergence with Multi-Projection Averaging: Through our novel multi-projection averaging mechanism with dynamic projections, \texttt{FedMPDD} achieves a convergence rate of $\mathcal{O}(1/\sqrt{K})$ (Theorem~\ref{thm:main_2}), comparable to the uncompressed \texttt{FedSGD} baseline. The parameter $m$ provides a tunable trade-off between communication cost and convergence speed.
    
    \item Inherent Privacy Preservation: As a natural by-product of our rank-deficient communication scheme, \texttt{FedMPDD} provides an inherent defense against GIA attacks without the need for convergence-degrading noise injection.
\end{itemize}

The remainder of this paper is organized as follows. Section~\ref{sec:mpdd_part} introduces the projected directional derivative, provides the motivation behind the proposed approach, and describes a naive implementation of the proposed mechanism and its challenges. Section~\ref{sec::Multi-projection} then presents the \texttt{FedMPDD} algorithm, details its update steps, and develops the convergence analysis, including convergence rates and privacy-preservation properties. Section~\ref{sec:experiment} reports numerical experiments, and Section~\ref{sec:conclo} concludes the paper. For brevity, all proofs and supplementary experimental results are presented in the appendices.

\section{{FL via Projected Directional Derivatives}}\label{sec:mpdd_part}
The \emph{projected directional derivative} is formally defined as:
\noindent\begin{definition}[\emph{projected directional derivative}]\label{def:proj}\rm{
For a differentiable function \( f : \mathbb{R}^d \to \mathbb{R} \) the \emph{projected directional derivative} is 
$\widehat{\nabla f}(\vect{x}) := \left( \vect{u}^\top \nabla f(\vect{x}) \right) \vect{u}$,
where $\nabla f(\vect{x})$ is the gradient of $f$ and \( \vect{u} \in \mathbb{R}^d \) is a random perturbation vector with entries \( u_i \) that are independently and identically distributed (i.i.d.) with zero mean and unit~variance.}\boxend
\end{definition}

\smallskip
The projected directional derivative  satisfies $\widehat{\nabla f}(\vect{x})^\top \nabla f(\vect{x}) \geq 0$ and $\mathbb{E}[\widehat{\nabla f}(\vect{x})] = \nabla f(\vect{x})$ (unbiased estimator for gradient) and thus makes $\vect{x}_{k+1} = \vect{x}_k - \eta \,\widehat{\nabla f}(\vect{x})$ to behave as a iterative successive descent (i.e., $\mathbb{E}[f(\vect{x}_{k+1})|\vect{x}_k] \approx f(\vect{x}_k) - \eta \| \nabla f(\vect{x})\|^2 < f(\vect{x}_k) \quad \text{for small } \eta > 0$). This stands in contrast to structured/sketched updates, whose gradient estimator $\widetilde{\nabla f}(\vect{x})$ is often biased ($\mathbb{E}[\widetilde{\nabla f}(\vect{x})] \neq \nabla f(\vect{x})$) and can violate the descent condition, thus lacking a general guarantee of~progress. 

\medskip
Within the framework of \texttt{FedSGD}, we consider an implementation that instead of the stochastic gradient $\vect{g}(\vect{x}_k)$, we employ the \emph{projected directional derivative} $\hat{\vect{g}}(\vect{x}_k)$ defined as: 
{\small
\begin{align}\label{eq:idea_proj}
    \hat{\vect{g}}(\vect{x}_k)
    &= \frac{1}{\beta N}\Bigg( 
        \sum_{i \in \mathcal{A}_k} 
        \underbrace{\vect{u}_{k,i}^\top \vect{g}_i(\vect{x}_k)}_{s_i^k:\,\textbf{client upload}}
        ~~\underbrace{\vect{u}_{k,i}}_{\textbf{server-side projection}}
    \Bigg) \nonumber \\
    &= \frac{1}{\beta N}\sum\nolimits_{i\in\mathcal{A}_k} s_i^k\,\vect{u}_{k,i}.
\end{align}
}

\begin{algorithm}[t]

\caption{\small \texttt{FedPDD}: \textbf{Fed}erated Learning via 
\textbf{P}rojected \textbf{D}irectional \textbf{D}erivative}
\label{alg:1}
\begin{algorithmic}[1]
\footnotesize
\State \textbf{Input:} $\vect{x}_0\!\in\!\mathbb R^{d}$, learning rate $\eta$,
      rounds $K$, client fraction $\beta\!\in(0,1]$
\For{each round \( k = 0, 1, \ldots, K-1 \)}
    \State Server samples client set $\mathcal A_k$ with size $\beta N$
    \State Server broadcasts $\vect{x}_k$ to all $i \in \mathcal{A}$
    \For{each client \( i \in \mathcal{A}_k \) \textbf{in parallel}}
        \State Generate i.i.d. Rademacher vector \( \vect{u}_k \in \{-1, +1\}^d \) using seed \( r_{k,i} \)
        \State Compute local stochastic gradient \( \vect{g}_i(\vect{x}_k) \)
        \State \textbf{Encode:} \( s_i^k = \vect{u}_{k,i}^\top \vect{g}_i(\vect{x}_k) \)
        \State Upload \( s_i^k \in \mathbb{R} \) and \( r_{k,i} \in \mathbb{R} \) to the server
    \EndFor
    \vspace{0.1in}
    \State $\vect{\Delta}_{\text{sum}} \gets \vect{0}_d$  \Comment{reset the estimator}
    
    \For{each client \( i \in \mathcal{A}_k \)} \Comment{on the server side}
    \vspace{0.1in}
    \State Server generates Rademacher vector \( \vect{u}_{k,i} \in \{-1, +1\}^d \) using seed \( r_{k,i} \) 

    \vspace{0.035in}
    \State \textbf{Decode:}\;   $\vect{\Delta}_{\text{sum}} 
         \;\gets\;
         \vect{\Delta}_{\text{sum}}  
         + 
         s_i^k\,\vect{u}_{k,i}^{}$
 \EndFor
  
    \State \textbf{Aggregate:}\; $ \displaystyle \hat{\vect g}(\vect{x}_k) = \frac{1}{\beta N}  \vect{\Delta}_{\text{sum}} $
    \State \textbf{Model update:}\; \( \vect{x}_{k+1} = \vect{x}_k - \eta \hat{\vect{g}}(\vect{x}_k) \)
\EndFor
\State \textbf{Output:} \( \vect{x}_K \)
\end{algorithmic}

\end{algorithm}

Considering the decomposition shown in~\eqref{eq:idea_proj}, we propose \texttt{FedPDD}, presented in Algorithm \ref{alg:1}, that proceeds as follows: each iteration begins with the server broadcasting the current model $\vect{x}_k \in \mathbb{R}^d$ to a sampled client set $\mathcal{A}_k$ (line~4). Upon receipt, each client generates a local random vector $\vect{u}_{k,i}$ using its private seed $r_{k,i}$ (line~6). The client encodes its local stochastic gradient into a scalar $s_{k,i}$ using the directional derivative along $\vect{u}_{k,i}$, and uploads this scalar together with the seed $r_{k,i}$, which is essential for enabling convergence (line~9). The server aggregates and decodes these scalars to update the global model $\vect{x}_{k+1}$ (lines 12-17). The design of \texttt{FedPDD} incorporates two key strategic choices. First, it employs a scalar seed $r_{k,i}$ to generate the identical $d$-dimensional vector $\vect{u}_{k,i}$ on the server side, thereby eliminating the need to transmit $\vect{u}_{k,i}$ directly while still ensuring convergence. Second, although any zero-mean, unit-variance distribution for $\vect{u}_{k,i}$ guarantees an unbiased projected directional derivative, our strategic choice of the Rademacher distribution for generating $\vect{u}_{k,i}$ yields lower variance compared to the standard normal distribution as established in the result below.
\begin{lem}[Variance Reduction via Distribution Choice; proof in Appendix~\ref{sec:proofs}]\label{var::reduction:dist_choice}
Consider the projected directional stochastic gradient $\hat{\vect{g}}(\vect{x})= \vect{u}^\top \vect{g}(\vect{x})\vect{u}$,   where the random direction $\vect{u} \in \mathbb{R}^d$ is drawn from either (i) the standard normal distribution $\mathcal{N}(\vect{0}, \mathbf{I}_d)$, or (ii) the Rademacher distribution with i.i.d.\ entries in $\{-1, +1\}$ (i.e., $\mathbb{P}(u_{j} = \pm 1) = \tfrac{1}{2}$ for all $j = 1, \dots, d$). The difference in variance between these two choices satisfies:
\begin{align*}
    \text{Var}_{\vect{u} \sim \mathcal{N}(\vect{0}, \mathbf{I}_d)}\left[\hat{\vect{g}}(\vect{x})\right]
    - \text{Var}_{\vect{u} \sim \text{Rademacher}^d}\left[\hat{\vect{g}}(\vect{x})\right]
    = 2\left\| \vect{g}(\vect{x}) \right\|^2 \mathbf{I}_d,
\end{align*}
where $\vect{g}(\vect{x})$ is the true stochastic gradient. \boxend
\end{lem}

\smallskip
The preliminary \texttt{FedPDD} algorithm offers two appealing properties. First, it provides significant communication reduction, as clients only transmit two scalars instead of a full gradient vector. Second, it ensures intrinsic privacy preservation due to the rank-deficient nature of the single-vector projection:
\begin{equation}\label{eq:rank1_property}
\underbrace{\hat{\vect{g}}_i(\vect{x}_k)}_{\textbf{Known}} = (\vect{u}_{k,i}^\top \vect{g}_i(\vect{x}_k)) \vect{u}_{k,i} = \underbrace{(\vect{u}_{k,i} \vect{u}_{k,i}^\top)}_{\textbf{Known, Rank-1 Matrix}} \!\!\!\vect{g}_i(\vect{x}_k).
\end{equation}
Since $\vect{u}_{k,i} \vect{u}_{k,i}^\top$ is a rank-1 matrix with a $(d-1)$-dimensional nullspace (by Sylvester's theorem~\cite{horn2012matrix}), equation~\eqref{eq:rank1_property} does not admit a unique solution: any $\vect{g}_i(\vect{x}_k)+\vect{v}$ with $\vect{v} \in \ker(\vect{u}_{k,i} \vect{u}_{k,i}^\top)$ also satisfies the equation. This large nullspace inherently introduces substantial uncertainty, making it impossible to uniquely reconstruct the true local gradient (see Section~\ref{sec::privacy} for further discussion). However, as the result below shows, these benefits are offset by poor convergence~performance. To analyze the convergence behavior of \texttt{FedPDD}, we first introduce some widely used conventional assumptions~\cite{t2020personalized, liu2022privacy, rostami2023federated} on the cost function, similar to those used in the analysis of FedSGD. 

\begin{assump}[Regularity Conditions on Local Objectives]\label{assump:smooth}
\rm{For each client \( i \), the local function \( f_i \) is \( L \)-smooth, i.e., for all \( \vect{x}, \vect{y} \in \mathbb{R}^d \),
$
\left\| \nabla f_i(\vect{x}) - \nabla f_i(\vect{y}) \right\| \leq L \left\| \vect{x} - \vect{y} \right\|.
$
In addition, the stochastic gradient is bounded in second moment: \( \mathbb{E}\big[\left\| \vect{g}_i(\vect{x}) \right\|^2\big] \leq G^2 \) for all \( \vect{x} \in \mathbb{R}^d \), and the variance of local gradients relative to the global gradient is bounded:
$
\frac{1}{N} \sum_{i=1}^{N} \left\| \nabla f_i(\vect{x}) - \nabla f(\vect{x}) \right\|^2 \leq \sigma^2$.
}
\boxend
\end{assump}

\begin{thm}[Convergence Bound of \texttt{FedPDD} Algorithm]\label{thm:main}
Consider \texttt{FedPDD} algorithm with a step size \( \eta = \frac{1}{L\sqrt{K}} \), and suppose that Assumption~\ref{assump:smooth} holds. 
Then, \texttt{FedPDD} algorithm converges to a stationary point of problem~\eqref{eq::opt_prob} at a rate of \( \mathcal{O}(d/\sqrt{K}) \), satisfying the following upper bound
{\small\begin{align}\label{eq:result_thm1_annotated}
\frac{1}{K} \sum\nolimits_{k=0}^{K-1} &\mathbb{E}\left[\left\| \nabla f(\vect{x}_k) \right\|^2\right]
\leq
\underbrace{\mathcal{O}\left( \frac{L(f(\vect{x}_0) - f^\star)}{\sqrt{K}}\right)}_{\text{due to initialization}}
+\nonumber \\
&\underbrace{\mathcal{O}\left( \frac{\sigma^2 (1/\beta -1)}{K\sqrt{K}} \right)}_{\text{due to client sampling}}~ + \!\!\!\!\!\!\!\!\underbrace{\mathcal{O}\left( \frac{d G^2}{\sqrt{K}} \right)\,,}_{\text{due to projected directional derivative}}\!\!\!\!
\end{align}}\normalsize
where $\beta\in (0, 1]$ denotes the client participation fraction, and \( f^\star \) denotes the global minimum of \( f \).\boxend
\end{thm}
\smallskip

The third term in the convergence bound~\eqref{eq:result_thm1_annotated} stems from the rank-1 projection operator $\vect{u}_{k,i} \vect{u}_{k,i}^\top$ (proof in Appendix \ref{sec:proofs}). Despite the projected directional derivative being an unbiased estimator, this rank-one map introduces high variance due to a magnitude scaling of $\sqrt{d}$ compared to the gradient:
{\small\begin{align*}
\mathbb{E}_{\vect{u}}\left[\bigl\|\hat{\mathbf g}_i(\mathbf x_k)\bigr\|\right]&\leq \sqrt{\mathbb{E}_{\vect{u}}\!\left[\bigl\|\hat{\mathbf g}_i(\mathbf x_k)\bigr\|^2\right]}\\&
=\sqrt{d\|\mathbf g_i(\mathbf x_k)\|^2}
=\sqrt{d}\,\bigl\|\mathbf g_i(\mathbf x_k)\bigr\|.
\end{align*}}
The uncontrollable $\sqrt{d}$ scaling yields the $\mathcal{O}(d/\sqrt{K})$ convergence rate highlighted in~\eqref{eq:result_thm1_annotated}, which scales poorly with model dimension $d$. Consequently, \texttt{FedPDD} requires roughly $d$ times more iterations than \texttt{FedSGD} ($\mathcal{O}(1/\sqrt{K})$ rate) to achieve comparable performance, resulting in total uplink communication of $\mathcal{O}(d/\sqrt{K} \times \beta N)$ bits asymptotically matching \texttt{FedSGD}'s cost and negating the per-round savings. While adjusting the step size to $\eta = \frac{1}{dL\sqrt{K}}$ could mitigate the $d$ factor in the projection-induced term, it necessitates a smaller overall step size that significantly slows convergence by amplifying the influence of the other terms in~\eqref{eq:result_thm1_annotated}. The larger variance also causes potential overshooting in the \texttt{FedPDD} updates. These critical limitations motivate the development of \texttt{FedMPDD} in Section~\ref{sec::Multi-projection} and Algorithm~\ref{alg:2}, which addresses these shortcomings through a multi-projection approach for practical large-scale problems.

\medskip
\section{{FL via Multi–Projected Directional Derivatives}}\label{sec::Multi-projection}
To address the limitation of \texttt{FedPDD}, we extend the estimator by sampling multiple directions. Specifically, at iteration \(k\), each selected client \(i\) draws \(m\) i.i.d.\ Rademacher vectors
\(\{\mathbf u^{(j)}_{k,i}\}_{j=1}^{m} \subset \{-1,+1\}^{d}\). To better understand the mechanism, we now stack the sampled vectors as columns to form the matrix \( \mathbf{U}_{k,i} \in \mathbb{R}^{d \times m} \).
Using this construction, the generalized estimator is defined as:
{\small
\begin{align}\label{eq:: m-projected-gradient}
\hat{\vect{g}}_i(\vect{x}_k)
   &:= \frac{1}{m}\sum_{j=1}^m 
      \bigl(\vect{u}^{(j)}_{k,i}{}^\top\vect{g}_i(\vect{x}_k)\bigr)\,\vect{u}^{(j)}_{k,i} = \frac{1}{m}\,\mathbf{U}_{k,i}\!\left(\mathbf{U}_{k,i}^\top \vect{g}_i(\vect{x}_k)\right), \nonumber\\
\mathbf{U}_{k,i} 
   &= \begin{bmatrix}
        \vect{u}_{k,i}^{(1)} & \vect{u}_{k,i}^{(2)} & \cdots & \vect{u}_{k,i}^{(m)}
      \end{bmatrix}
      \in \mathbb{R}^{d\times m}.
\end{align}
}
As \( \mathbb{E}[\mathbf{U}_{k,i} \mathbf{U}_{k,i}^{\!\top}] = m \mathbf{I}_d \), the estimator remains unbiased:
$
\mathbb{E}\bigl[\hat{\vect{g}}_{i}(\vect{x}_k)\bigr]
= \frac{1}{m} \,\mathbb{E}[\mathbf{U}_{k,i} \mathbf{U}_{k,i}^{\!\top}] \, \vect{g}_{i}(\vect{x}_k)
   = \vect{g}_{i}(\vect{x}_k).$

By constructing, the mapping \( \tfrac{1}{m} \mathbf{U}_{k,i} \mathbf{U}_{k,i}^\top \) satisfies the high-probability operator-norm Johnson–Lindenstrauss (JL) Lemma \cite{vershynin2018high}.
\begin{lem}[JL Bound for Multi-Projected Directional Derivatives]\label{lem:opJL}
For any \(0<\varepsilon<1\) and \(0<\delta<1\), if the number of sampled random directions satisfies
\begin{equation}\label{eq::m_choice}
m = \mathcal{O}\left( \frac{\ln(d/\delta)}{\varepsilon^{2}} \right),
\end{equation}
then with probability at least \( 1 - \delta \), the following bound holds:
\begin{equation}\label{jl_bound}
\left\| \frac{1}{m} \mathbf{U}_{k,i} \left( \mathbf{U}_{k,i}^\top \vect{g}_i(\vect{x}_k) \right) \right\| 
\leq (1 + \varepsilon) \left\| \vect{g}_i(\vect{x}_k) \right\|,
\end{equation}
where $\mathbf{U}_{k,i}$ is defined in~\eqref{eq:: m-projected-gradient} and its columns are sampled independently according to the projection direction distribution in Definition~\ref{def:proj}.\boxend
\end{lem}

\medskip

The proof of this lemma follows trivially from the JL Lemma~\cite[Theorem 5.3]{vershynin2018high}. 
This result implies that the mapping operator \( \tfrac{1}{m} \mathbf{U}_{k,i} \mathbf{U}_{k,i}^\top \) approximately preserves the norm of the client's gradient with high probability, provided a sufficient number of sampled directions. Moreover, as \( m \to \infty \), the mapping approaches the identity operator \( \tfrac{1}{m} \mathbf{U}_{k,i} \mathbf{U}_{k,i}^\top \to \mathbf{I}_d \) in expectation, due to the unbiasedness of the projection. Motivated by this probabilistic guarantee, which grows only logarithmically with the ambient dimension \( d \), we design \texttt{FedMPDD} algorithm, presented in Algorithm \ref{alg:2}, as a generalization of \texttt{FedPDD} algorithm. The following result provides a convergence guarantee for \texttt{FedMPDD}.

\begin{thm}[Convergence Bound of \texttt{FedMPDD} Algorithm]\label{thm:main_2}
Let the step size \( \eta = \frac{1}{L\sqrt{K}} \), and suppose that Assumption~\ref{assump:smooth} holds. Let number of random vectors be $m=O \bigl(\tfrac{\ln(d/\delta)}{\varepsilon^{2}}\bigr)$. Then, \texttt{FedMPDD} algorithm converges to a stationary point of problem~\eqref{eq::opt_prob} at a rate of \( \mathcal{O}(1/\sqrt{K}) \), satisfying the following upper bound with probability at least $1-\delta$,
{\small\begin{align}\label{eq:result_thm1_annotated_thm_2}
\frac{1}{K} \sum\nolimits_{k=0}^{K-1} &\mathbb{E}\left[\left\| \nabla f(\vect{x}_k) \right\|^2\right]
\!\leq
\!\underbrace{\mathcal{O}\!\left( \frac{L(f(\vect{x}_0) - f^\star)}{\sqrt{K}} \right)}_{\text{due to initialization}} +\nonumber \\
 &\underbrace{\mathcal{O}\!\left( \frac{\sigma^2(1/\beta -1)}{K\sqrt{K}} \right)}_{\text{\!\!\!\!\!\!\!\!due to client sampling}} ~+ \!\!\!\!\!\!\!\!\!\!\underbrace{\!\!\mathcal{O}\left( \frac{\epsilon G^2}{\sqrt{K}} \right)}_{\text{due to Multi-projected directional derivatives}}\!\!\!\!\!\!\!\!\!\!\!\!\!\!\!\!\!\!\!\!\!\!\!\!\!,
\end{align}}
where $0< \epsilon < 1$ is the distortion parameter, $\beta\in (0, 1]$ denotes the client participation fraction, and \( f^\star \) denotes the global minimum of \( f \).\boxend
\end{thm}

\begin{remark}\label{remak:choosing_m}
[Choice of $m$, $\varepsilon$, and $\delta$]\rm{
The number of random directions $m$ is selected according to the JL embedding requirement in~\eqref{eq::m_choice} 
where $\varepsilon\in(0,1)$ controls the distortion of the projected gradient and $\delta\in(0,1)$ is the failure probability.
In our setting, $\varepsilon$ directly appears in the convergence bound~\eqref{eq:result_thm1_annotated_thm_2} of Theorem~\ref{thm:main_2} through the term
$\mathcal{O}\!\left(\frac{\varepsilon G^2}{\sqrt{K}}\right)$, and therefore acts as an accuracy knob, while $\delta$ determines the confidence level with which the JL property holds. In practice, we fix $\delta$ to a small constant (e.g., $\delta\in[10^{-3},10^{-1}]$) to ensure high-probability guarantees, and select $\varepsilon$ in a moderate distortion regime (e.g., $\varepsilon\in[0.05,0.2]$). With these choices, $m$ is then computed using the above scaling and treated as a communication--accuracy trade-off parameter.
In our experiments, we set $m$ to be a small fraction of the ambient dimension, specifically in the range $m \in [0.2\%,\,4\%]\times d$, which lies well within the JL-valid regime predicted by theory. All experimental results are conducted within this regime, ensuring that the theoretical guarantees apply throughout training.\boxend
}
\end{remark}

\begin{algorithm}[t]
{\scriptsize
\caption{ \texttt{FedMPDD}: \textbf{Fed}erated Learning via \textbf{M}ulti-\textbf{P}rojected \textbf{D}irectional \textbf{D}erivatives}
\label{alg:2}
\makeatletter
\renewcommand{\alglinenumber}[1]{\scriptsize #1:}
\makeatother

\begin{algorithmic}[1]

\State \textbf{Input:}  $\vect{x}_0\!\in\!\mathbb R^{d}$, learning rate $\eta$,
      rounds $K$, $\#~\text{random directions} ~~m$, client fraction $\beta\!\in(0,1]$
\For{$k = 0,1,\dots,K-1$}
    \State Server samples client set $\mathcal A_k$ with $|\mathcal A_k|= \beta N$
    \State Server broadcasts $\vect{x}_k$ to all $i\in\mathcal A_k$
    \For{\textbf{each} client $i\in\mathcal A_k$ \textbf{in parallel}}

    \State Compute local stochastic gradient \( \vect{g}_i(\vect{x}_k) \)
    \For{$j = 1, \dots, m$}   \Comment{loop over projected directions}
        \State Client generates i.i.d.\ Rademacher vector \( \vect{u}_{k,i}^{(j)} \in \{-1, +1\}^d \) using seed \( r_{k,i} \) and index \( j \)
        \State \textbf{Encode:} \( \vect{s}_i^k[j] \gets \bigl(\vect{u}_{k,i}^{(j)}\bigr)^\top \vect{g}_i(\vect{x}_k) \)
    \EndFor
    
    \State Upload $\vect {s}_i^k \in\mathbb R^{m} $ and $r_{k,i} \in\mathbb R $ to the server
\EndFor

\State $\vect{\Delta}_{\text{sum}} \gets \vect{0}_d$  \Comment{reset the estimator}
 \For{each client \( i \in \mathcal{A}_k \)} \Comment{on the server side}
    
    \For{$j = 1, \dots, m$}   \Comment{re-generate the same projected directions}
    \State Server generates i.i.d.\ Rademacher vector $\vect{u}_{k,i}^{(j)} \in \{-1,+1\}^d$ using seed $r_{k,i}$ and index $j$
    \vspace{0.025in}
    \State \textbf{Decode:}\;   $\vect{\Delta}_{\text{sum}} 
         \;\gets\;
         \vect{\Delta}_{\text{sum}}  
         + 
         \dfrac{\vect {s}_i^k[j]}{m}\,\vect{u}_{k,i}^{(j)}$
    \EndFor
\EndFor

      \State \textbf{Aggregate:}\;
           $ \displaystyle \hat{\vect g}(\vect{x}_k)
               =\frac{1}{\beta N}  \vect{\Delta}_{\text{sum}} $
               
    \State \textbf{Model update:}\;
           $\vect{x}_{k+1} = \vect{x}_k - \eta\,\hat{\vect g}(\vect{x}_k)$
\EndFor
\State \textbf{Output:} $\vect{x}_K$
\end{algorithmic}
}
\end{algorithm}

\begin{remark}[Computational Cost of \texttt{FedMPDD}]\label{remark:broader_impact}
\rm{
The client-side encoding in \texttt{FedMPDD} has a computational cost of 
$\mathcal{O}(dm)$ (see lines~7--10 of the \texttt{FedMPDD} algorithm; as reported in 
Table~\ref{tab:mpdd_latency} for one representative experiment, this 
computational time is negligible and does not constitute a bottleneck in our 
experiments). While this may initially seem costly, it is often offset in practice, since in many federated learning settings, client models are deep neural networks and computing the full stochastic gradient (line~6) is already expensive. Recent work~\cite{baydin2022gradients, ren2022scaling, silver2021learning} has shown that computing the inner product \( \vect{u}^\top \vect{g}_i \) is significantly more efficient than computing the full gradient \( \vect{g}_i \), because the operation can be implemented as a Jacobian-vector product (JVP), which leverages efficient vector-matrix multiplication in deep networks. Specifically, projected-forward methods reduce the time complexity of gradient computation from \( \mathcal{O}(h^2 p T^2) \) (for full forward-mode autodiff) to \( \mathcal{O}(h^2 T + hpT) \), where \( h \), \( p \), and \( T \) denote the hidden dimension, the number of parameters per layer, and the total number of layers, respectively. Motivated by these insights, \texttt{FedMPDD} can avoid computing \( \vect{g}_i \) explicitly (line~6) by fixing a single mini-batch \( \mathcal{B}_i^k \) and reusing it across all random directions. The encoding step (line~9) is then performed via the projected-forward approach using JVPs. We can show that when \( m < \tfrac{hpT}{h + p} \), this strategy reduces overall client-side computation, making \texttt{FedMPDD} particularly suitable for resource-constrained devices. We empirically evaluate this strategy in our follow-up study (see Section~\ref{sec:experiment}). 
}
\boxend.
\end{remark}

\begin{remark}[Communication Reduction and Efficiency in \texttt{FedMPDD}]
\rm{
\texttt{FedMPDD} presented in Algorithm \ref{alg:2} significantly reduces per-round uplink communication by enabling clients to transmit only an $m$-dimensional vector $\vect{s}_i^k \in \mathbb{R}^m$ together with a scalar seed number $r_{k,i} \in \mathbb{R}$, instead of the full $d$-dimensional gradient ($m \ll d$). This is achieved by encoding the client's $d$-dimensional gradient through its projection onto a set of $m$ random scalars (line 9). Moreover, the total communication cost over the full training horizon is reduced to $\mathcal{O}(1/\sqrt{K} ~\times \beta N \times m)$, where \( \beta \) is the client participation ratio and \( N \) is the number of clients. Since \( m \) grows only logarithmically with the problem dimension \( d \), the communication savings become even more substantial for large-scale models. Our experiments such as one with model dimension $d=319{,}242$, show that using only $m=600$ ($\approx 0.2\%$ of $d$) directions, \texttt{FedMPDD} matches the accuracy and convergence speed of baseline methods while substantially reducing uplink communication.
}\boxend
\end{remark}

\subsection{Privacy Preservation Attributes of \texttt{FedMPDD}}\label{sec::privacy}

The privacy guarantees of \texttt{FedMPDD} are demonstrated under a standard honest-but-curious threat model.

\begin{definition}[\emph{Threat model}]\label{threat_model}\rm{
An \emph{honest-but-curious} adversary (e.g., the server) correctly follows the protocol but attempts to infer private client data by analyzing all accessible information, which includes communication messages, model architecture, and global hyperparameters.
}\end{definition}

\medskip\noindent
Against this adversary, \texttt{FedMPDD}'s privacy stems from its rank-deficient projection ($m \ll d$), which creates quantifiable uncertainty for any party observing the transmitted data. We formalize this protection below.

\begin{lem}[Gradient reconstruction error; proof in Appendix~\ref{sec:proofs}]\label{lem:var_estimator_mpdd_final_adjusted}
For the reconstructed gradient estimator $\hat{\vect{g}}_i(\vect{x}_k) = \frac{1}{m} \sum_{j=1}^m \vect{u}^{(j)}_{k,i} (\vect{u}^{(j)}_{k,i})^\top \vect{g}_i(\vect{x}_k)$, the expected relative squared error is:
\begin{equation}\label{eq:relative_error_mpdd_final_adjusted}
\mathbb{E}_{\mathbf{U}} \big[ \| \hat{\vect{g}}_i(\vect{x}_k) - \vect{g}_i(\vect{x}_k) \|^2 \big] \big/ \| \vect{g}_i(\vect{x}_k) \|^2 = \frac{d-1}{m}.
\end{equation}\boxend
\end{lem}
\medskip

This inherent gradient ambiguity provides a formal defense against GIA attacks by establishing a lower bound on reconstruction error. 

The following Lemma establishes a lower bound on the private data reconstruction
error. We denote by $(v,c)$ a private training sample held by a client, where $v$
represents the raw input data (e.g., an image) and $c$ denotes the corresponding label.

\begin{lem}[Lower bound on private data reconstruction error; proof in Appendix~\ref{sec:proofs}]\label{lem:feature_recovery_bound}
Suppose an adversary attempts to reconstruct a private input vector $v$ by minimizing $\mathcal{L}(\hat{v}) := \| \tfrac{1}{m} \vect{U}_{k,i} \vect{U}_{k,i}^\top \vect{g}_i(v,c;\vect{x}_k) - \vect{g}_i(\hat{v},c;\vect{x}_k) \|$. The expected reconstruction error is lower bounded by:
\begin{equation}\label{eq:feature_recovery_bound}
\mathbb{E}\big[ \| v - \hat{v}^* \|^2 \big] \geq \frac{d-1}{m \cdot L_v(\vect{x})^2} \| \vect{g}_i(v,c;\vect{x}_k) \|^2,
\end{equation}
where $L_v(\vect{x})$ is the Lipschitz constant of the gradient with respect to $v$.\boxend
\end{lem}

\medskip
Together, these lemmas establish that the gradient reconstruction error of $\frac{d-1}{m}$ translates into a concrete lower bound on data recovery, creating a privacy barrier that scales with model dimension $d$.

\medskip
Our approach offers some fundamental advantages over additive-noise methods. In FL with LDP, adding noise $\boldsymbol{\zeta}_i$ to each client's gradient yields a relative reconstruction error proportional to:
\small
\begin{align}
\label{eq::priv_reconstruc_LDP}
{\small \text{Relative Reconstruction Error}}\!\!
\;&\propto\; \!\!\!
\frac{
\mathbb{E}_{\zeta_i}\!\bigg[
\bigg\|
\overbrace{
\vect{g}_i(\vect{x}_k) + \boldsymbol{\zeta}_i
}^{\substack{\textbf{honest-but-curious}\\\textbf{server observation}}}
- \vect{g}_i(\vect{x}_k)
\bigg\|^2
\bigg]
}{\|\vect{g}_i(\vect{x}_k)\|^2}
\nonumber\\
&=\;
\frac{\mathbb{E}_{\zeta_i}\!\big[\|\boldsymbol{\zeta}_i\|^2\big]}{\|\vect{g}_i(\vect{x}_k)\|^2}
\;=\;
\frac{d\,\tau^2}{\|\vect{g}_i(\vect{x}_k)\|^2}.
\end{align}

where $\tau^2$ is the variance of noise $\zeta_i$. This creates an inconsistency: large gradients are poorly protected, while small gradients can be overwhelmed by noise, potentially breaking convergence (when $\|\boldsymbol{\zeta}\| \gg \|\vect{g}\|$, the perturbed gradient can flip the descent direction). Achieving consistent privacy with LDP requires large noise that degrades performance, and still incurs $O(d)$ communication cost. In contrast, \texttt{FedMPDD} provides a \emph{consistent} relative reconstruction error of $\frac{d-1}{m}$ that is independent of gradient magnitude. This design simultaneously ensures: (i) guaranteed descent direction ($\vect{g}^\top \hat{\vect{g}} \geq 0$) without added noise, (ii) constant privacy through a fixed $(d-m)$-dimensional nullspace, and (iii) substantial communication savings ($m \ll d$).

\begin{figure}[t]
\centering
    \includegraphics[width=0.43\textwidth]{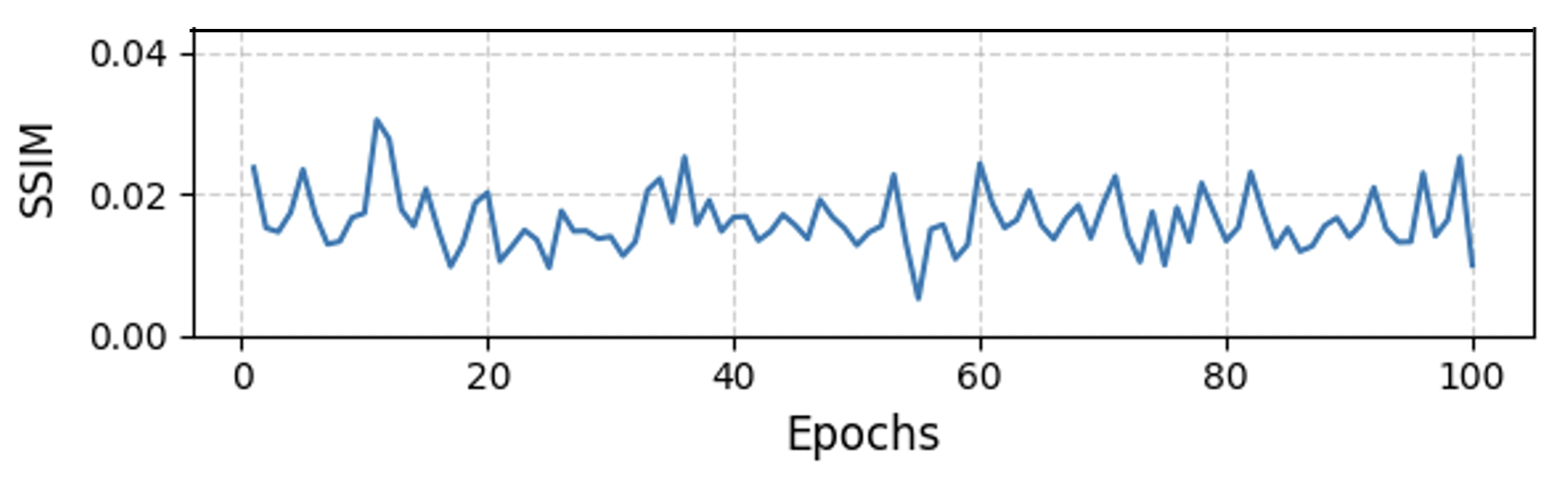}
    \caption{{\footnotesize SSIM scores from GIA~\cite{yu2025gi} on LeNet using \texttt{FedMPDD} with $m = 600$ remain consistently low (below $0.04$) over $100$ training epochs, demonstrating that privacy protection is independent of the training stage.
}}
\label{fig:ssim_lenet}
\end{figure}

\begin{figure*}[t]
\centering
\begin{minipage}[t]{0.37\linewidth}
    \centering
    \includegraphics[width=\linewidth]{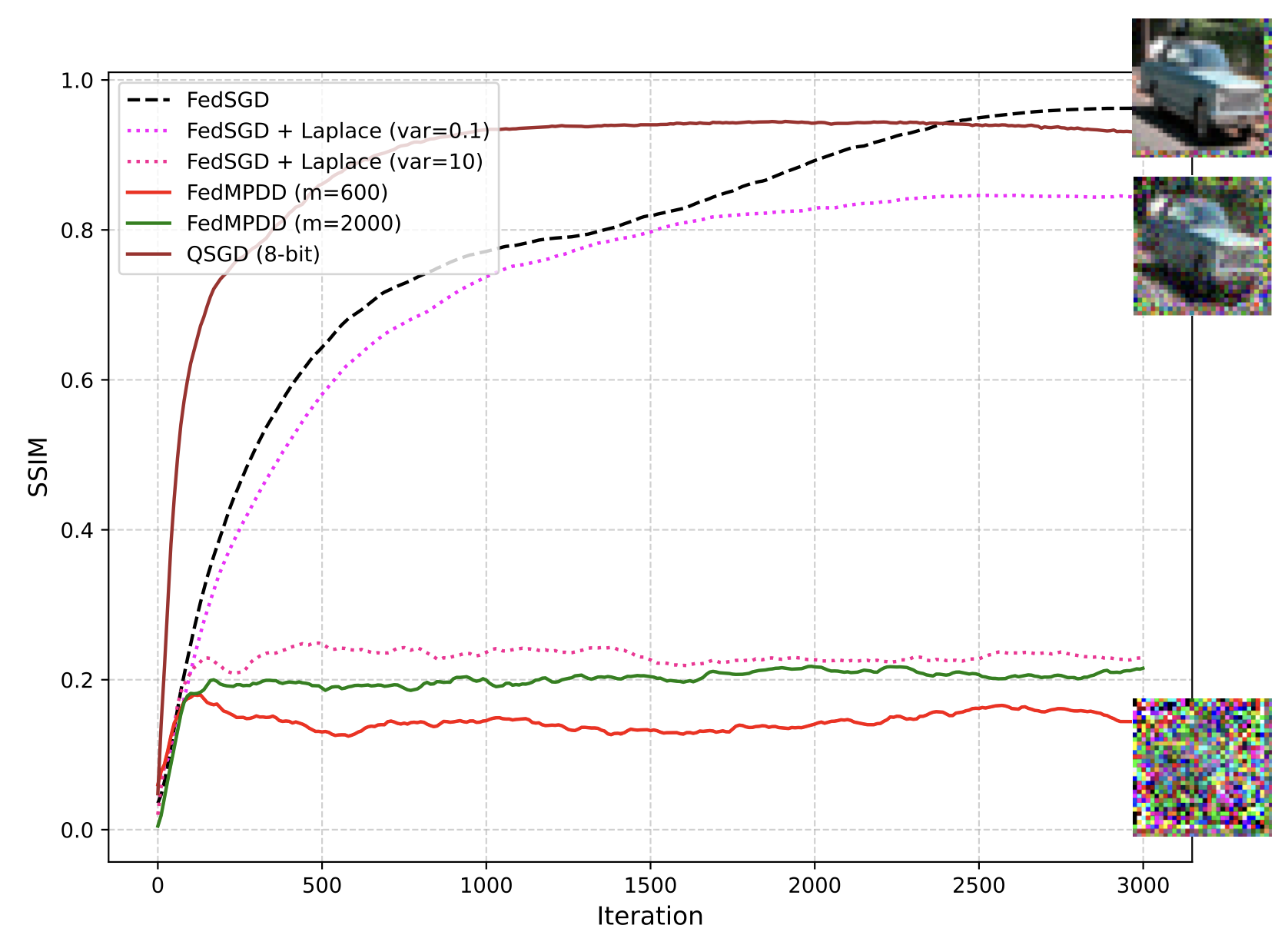}
\end{minipage}
\hfill
\begin{minipage}[t]{0.55\linewidth}
    \centering
    \includegraphics[width=\linewidth]{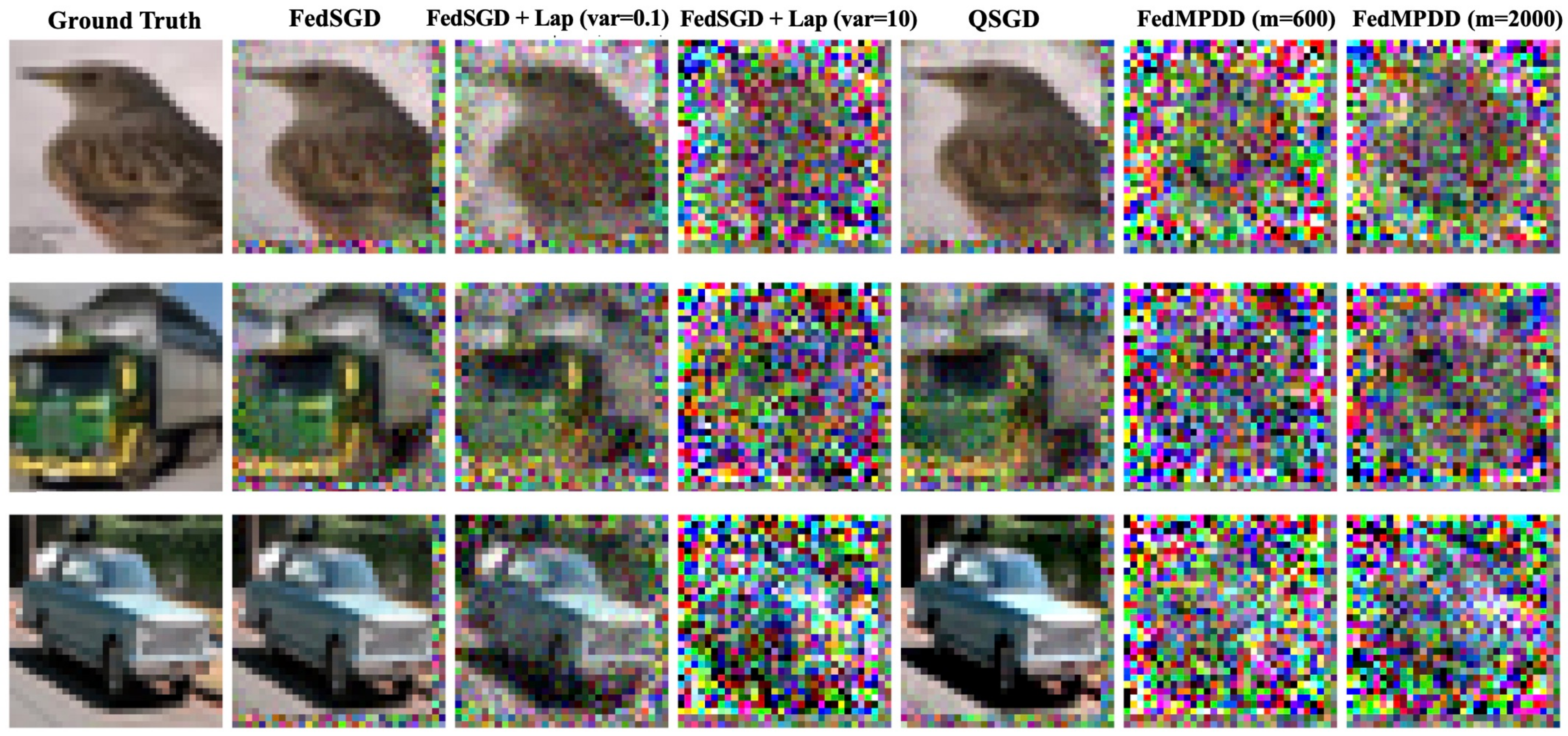}
\end{minipage}
\caption{{\footnotesize
GIA attack~\cite{yu2025gi} visualization: SSIM scores (left) and reconstructed CIFAR-10 samples (right). LDP with small noise (columns 3--4) and QSGD (column 5) show significant data leakage, while \texttt{FedMPDD} (columns 6--7) demonstrates stronger privacy.
}}
\label{cifa10_visul_attack}
\end{figure*}

Our theoretical guarantees are supported by empirical results. Figure~\ref{cifa10_visul_attack} illustrates \texttt{FedMPDD}'s data obfuscation on CIFAR-10 for different values of $m$. Figure~\ref{fig:ssim_lenet} shows that SSIM scores remain consistently low across training epochs, confirming that privacy protection is independent of gradient magnitude and training stage. However, a critical consideration remains when an adversary observes a client over multiple rounds. We formalize this concern below.

\begin{lem}[Worst-Case Multi-Round Privacy Bound; proof in Appendix~\ref{sec:proofs}]
\label{thm:worst_case_privacy}
Consider the worst-case scenario where the client's gradient $g_i$ remains constant across all rounds. An adversary observing projections from $T$ rounds obtains $T \times m$ linear constraints on the $d$-dimensional gradient. Privacy is preserved (gradient cannot be uniquely recovered) as long as $T < \frac{d}{m}$. \boxend
\end{lem}
\smallskip
The bout $T < \frac{d}{m}$ in this statement is due to the underdetermined system having a non-trivial nullspace of dimension $d - T \times m > 0$. This bound reflects an \emph{observability constraint}: the adversary's rank-$m$ measurements per round are insufficient to reconstruct the full $d$-dimensional gradient. With typical settings ($d \in [10^6, 10^9]$, $m = 100$, $T \in [10^2, 10^4]$), the condition $T \times m < d$ is easily satisfied (e.g., $d = 10^6$ and $m = 100$ allows $T < 10^4$ rounds). Moreover, gradient evolution during training provides additional protection beyond this worst-case static analysis~\cite{geiping2020inverting}.

\begin{remark}[Privacy-communication trade-off]\rm{ The parameter $m$ serves as a tunable knob for the privacy-communication-accuracy trade-off. From a system-theoretic viewpoint, $m$ controls the rank of the observation operator at each time step, regulating the observability of the client's gradient trajectory~\cite{teixeira2015secure}. A larger $m$ increases communication overhead (uplink message size scales with $m$) and reduces privacy (more gradient information revealed), but improves gradient reconstruction accuracy, with expected error $\tfrac{d-1}{m}$ (Lemmas~\ref{lem:var_estimator_mpdd_final_adjusted} and~\ref{lem:feature_recovery_bound}). This fundamental trade-off where improving one aspect (e.g., accuracy) degrades another (e.g., privacy or communication efficiency) mirrors similar tensions in differential privacy. The worst-case bound $T < d/m$ directly quantifies this: smaller $m$ allows more training rounds while maintaining privacy, but may require more rounds to achieve convergence.}\boxend
\end{remark}

\smallskip
Our experimental results across two attack families, including the recent GIA attacks~\cite{yu2025gi} and the well-known deep leakage from gradients method~\cite{zhu2019deep} (details in Appendix~\ref{DLG:explation}), support this theoretical finding and demonstrate that the predicted privacy protection holds in practice.

\section{Numerical Experiments}\label{sec:experiment}

We conduct a comprehensive evaluation of the proposed algorithm against the baseline methods using multiple neural network architectures with varying parameter sizes: the LeNet model with $d = 13{,}426$, the CNN architecture from~\cite{lin2022personalized} with $d = 61{,}706$, and the CNN model from~\cite{mcmahan2017communication} with $d = 319{,}242$. Unless stated otherwise, data is partitioned among $100$ clients; Tables~\ref{tab:lenet_arch}-\ref{tab:smallcnn_arch} gives the model architectures of these benchmark models. Note that, MNIST and FASHIONMNIST datasets each contain $60{,}000$ training samples and $10{,}000$ test samples. The CIFAR-10 dataset consists of $50{,}000$ training samples and $10{,}000$ test samples. Moreover, under the non-i.i.d. data distribution, each client receives data from exactly two classes in the dataset. Experiments involving logistic regression are conducted on a MacBook Pro CPU. All other experiments are executed on an NVIDIA A100 GPU.

\begin{table}[t]
\centering
\scriptsize
\caption{Model architecture of LeNet.}
\label{tab:lenet_arch}
\begin{tabular}{llccc}
\toprule
\textbf{Layer} & \textbf{Type} & \textbf{Kernel Size} & \textbf{Output Shape} & \textbf{Activation} \\
\midrule
Input     & -             & -         & $1 \times 28 \times 28$ & - \\
Conv1     & Conv2D        & $5 \times 5$, stride 2, padding 2 & $12 \times 14 \times 14$ & Sigmoid \\
Conv2     & Conv2D        & $5 \times 5$, stride 2, padding 2 & $12 \times 7 \times 7$  & Sigmoid \\
Conv3     & Conv2D        & $5 \times 5$, stride 1, padding 2 & $12 \times 7 \times 7$  & Sigmoid \\
Flatten   & -             & -         & $588$                   & - \\
FC        & Linear        & -         & $10$ (logits)           & - \\
\bottomrule
\end{tabular}
\end{table}

\begin{table}[t]
\centering
\scriptsize
\caption{Model Architecture of CNN in \cite{lin2022personalized}.}
\label{tab:smallcnn_arch}
\begin{tabular}{llccc}
\toprule
\textbf{Layer} & \textbf{Type} & \textbf{Kernel / Params} & \textbf{Output Shape} & \textbf{Activation} \\
\midrule
Input     & -          & -                  & $1 \times 28 \times 28$  & - \\
Conv1     & Conv2D     & $5 \times 5$, MaxPool(2) & $6 \times 12 \times 12$ & ReLU \\
Conv2     & Conv2D     & $5 \times 5$, MaxPool(2) & $16 \times 4 \times 4$  & ReLU \\
Flatten   & -          & -                  & $400$                    & - \\
FC1       & Linear     & $400 \rightarrow 120$    & $120$                   & ReLU \\
FC2       & Linear     & $120 \rightarrow 84$     & $84$                    & ReLU \\
FC3       & Linear     & $84 \rightarrow 10$      & $10$ (logits)           & - \\
\bottomrule
\end{tabular}
\end{table}

\begin{table}[t]
\centering
\scriptsize
\caption{Model Architecture of CNN in \cite{mcmahan2017communication}.}
\label{tab:cnn_cifar_arch}
\begin{tabular}{llccc}
\toprule
\textbf{Layer} & \textbf{Type} & \textbf{Kernel / Params} & \textbf{Output Shape} & \textbf{Activation} \\
\midrule
Input     & -          & -                  & $3 \times 32 \times 32$   & - \\
Conv1     & Conv2D     & $3 \times 3$       & $128 \times 30 \times 30$ & ReLU \\
          & MaxPool    & $2 \times 2$       & $128 \times 15 \times 15$ & - \\
Conv2     & Conv2D     & $3 \times 3$       & $128 \times 13 \times 13$ & ReLU \\
          & MaxPool    & $2 \times 2$       & $128 \times 6 \times 6$   & - \\
Conv3     & Conv2D     & $3 \times 3$       & $128 \times 4 \times 4$   & ReLU \\
Flatten   & -          & -                  & $2048$                    & - \\
FC1       & Linear     & $2048 \rightarrow 10$ & $10$ (logits)          & - \\
\bottomrule
\end{tabular}
\end{table}

For a comprehensive evaluation, we tested client participation rates of $10\%$, $50\%$, and $100\%$ across different tasks, considering both i.i.d. and non-i.i.d. data distributions (where each client accesses only a subset of classes in multi-class classification). Hyperparameter tuning details for each model are in Appendix~\ref{hyper_parameters}. For gradient inversion attacks, we employed two algorithms: (i) the recent method proposed by~\cite{yu2025gi}, and (ii) the well-known Deep Leakage from Gradients (DLG) algorithm~\cite{zhu2019deep}, which reconstructs original input data (e.g., images) from shared gradients in distributed learning.

We compare the communication cost reduction of \texttt{FedMPDD} against a recent sketching-based method~\cite{lin2022personalized}, a structured-based method~\cite{yang2024spd}, a top-$k$ sparsification method~\cite{alistarh2018convergence}, and the quantization-based method QSGD~\cite{alistarh2017qsgd}. For performance evaluation, FedSGD serves as the accuracy baseline. Our communication cost analysis includes total and per-round uplink overhead, as well as \emph{i)} performance under a constrained communication budget and \emph{ii)} the total communication cost to achieve target accuracy. To empirically validate \texttt{FedMPDD}'s privacy enhancement against GIAs, we compare it to LDP with varying noise levels in image classification tasks.

\begin{figure*}[t]
\centering    \includegraphics[scale=0.35]{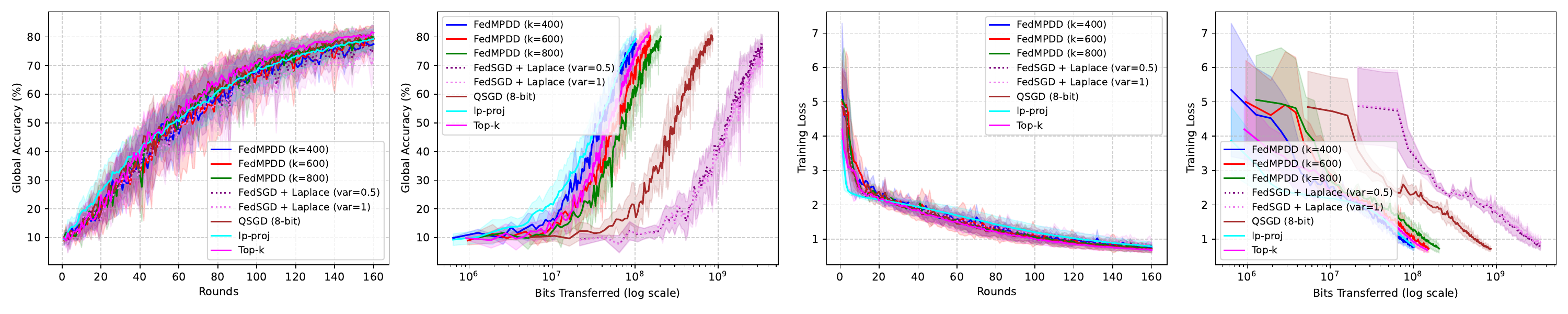}
    \caption{{\small Training loss and accuracy curves versus communication rounds and number of transmitted bits for the LeNet model on the MNIST dataset (i.i.d.).}}
    \label{fig}
\end{figure*}

\begin{table*}[t]
\centering
\caption[{Comparison of accuracy, communication, privacy, and reconstruction on MNIST}]%
{Comparison of test accuracy (under a fixed communication budget),
communication cost (under a target accuracy), privacy leakage, and
reconstruction quality using the attack of \protect\cite{yu2025gi}
on MNIST (i.i.d.) with LeNet.}

\scriptsize
\begin{tabular}{l c c|c c|c c}
\toprule
\textbf{Method} & \textbf{Bytes Budget (GB)} & \textbf{Test Acc (\%)} & \textbf{Target Acc (\%)} & \textbf{Used Bytes (GB)} & \textbf{Defendability} & \textbf{SSIM} \\
\midrule
FedSGD                          & 0.09 & 11.45 & 60 & 1.439 & \xmark & 1.00 \\
FedSGD + Laplace (var=0.5)      & 0.09 & 11.13 & 60 & 1.611 & \cmark & $\ll$ 0.03 \\
FedSGD + Laplace (var=1)        & 0.09 & 11.41 & 60 & 1.869 & \cmark & $\ll$ 0.03 \\
\textbf{FedMPDD (m=400, 2\% d)} & 0.09 & \textbf{77.37} & 60 & \textbf{0.052} & \cmark & $\ll$ 0.03 \\
\textbf{FedMPDD (m=600, 3\% d)} & 0.09 & \textbf{67.75} & 60 & \textbf{0.079} & \cmark & $\ll$ 0.03 \\
\textbf{FedMPDD (m=800, 4\% d)} & 0.09 & \textbf{58.49} & 60 & \textbf{0.093} & \cmark & $\ll$ 0.03 \\
QSGD (8-bit)                    & 0.09 & 21.66 & 60 & 0.376 & \xmark & 0.98 \\
Top-k (k=400)   & 0.09 & 65.75 & 60 & 0.077 & \xmark & 0.89 \\
lp-proj                         & 0.09 & 73.01 & 60 & 0.069 & \xmark & 0.75 \\
\bottomrule
\end{tabular}
\label{tab:fedmpdd_comparison_vrules_2}
\end{table*}

\begin{table*}[t]
\caption{{Comparison of test accuracy (under a fixed communication budget),
communication cost (under a target accuracy), privacy leakage, and reconstruction quality using the attack of \cite{yu2025gi} on CIFAR-10 (i.i.d.) with the CNN model from \cite{mcmahan2017communication}. The symbol $\star$ indicates that the communication budget was exceeded in the first iteration.}}

\centering

\label{cnn_fedavg_pp_table}
\scriptsize
\begin{tabular}{lcc|cr|cc}
\toprule
\textbf{Method} & \textbf{Bytes Budget (GB)} & \textbf{Test Acc (\%)} & \textbf{Target Acc (\%)} & \textbf{Used Bytes (GB)} & \textbf{Defendability} & \textbf{SSIM} \\
\midrule
FedSGD                              & 0.90 & $\star$          & 60 & 471.96 & \xmark & 0.96 \\
FedSGD + Laplace (var=0.1)          & 0.90 & $\star$          & 60 & 471.96 & \xmark & 0.84 \\
FedSGD + Laplace (var=10)           & 0.90 & $\star$          & 60 & not reached & \cmark & 0.23 \\
\textbf{FedMPDD (m=600, 0.2 \% d)}  & 0.90 & \textbf{40.84}   & 60 & \textbf{1.32} & \cmark & 0.14 \\
\textbf{FedMPDD (m=2000, 0.6 \% d)} & 0.90 & \textbf{36.26}   & 60 & \textbf{3.26} & \cmark & 0.22 \\
QSGD (8-bit)                        & 0.90 & 12.97            & 60 & 117.98 & \xmark & 0.93 \\
lp-proj                             & 0.90 & 34.72            & 60 & 1.84  & \xmark & 0.74 \\
Top-k (k=600)                       & 0.90 & 38.11            & 60 & 2.30  & \xmark & 0.91 \\
SA-FedLora                          & 0.90 & 35.84            & 60 & 2.10  & \xmark & 0.83 \\
\bottomrule
\end{tabular}%

\end{table*}

To evaluate the quality of reconstructed images after the attack, we employ the Structural Similarity Index Measure (SSIM)~\cite{lang2023joint}, a widely used metric for assessing image similarity, where an SSIM value closer to $1$ indicates a higher resemblance between the reconstructed image and the ground truth. 
For brevity, we only show a subset of the results and the full set of tables and training and accuracy curves are presented in Appendix \ref{additional_exp}. 
Note that in our experiments, we did not fine-tune the value of \( m \) to explicitly optimize for the minimal communication cost and maximal privacy guarantees achievable by \texttt{FedMPDD}. Instead, we selected $m = \mathcal{O}\left(\tfrac{\ln(d/\delta)}{\varepsilon^2}\right)$ for sufficiently small values of \( \delta \) and \( \varepsilon \). In Table~\ref{tab:fedmpdd_m_sweep} of Appendix~\ref{additional_exp}, we present experiments over a wide range of $m$ values to further illustrate the findings of Theorem~\ref{thm:main_2}. As $m$ becomes too small, both the rate of convergence and the accuracy deteriorate. Moreover, as we show in our reported results, the chosen values of \( m \) grow slightly with the parameter dimension \( d \) (ranging from a simple logistic model to a deep CNN model with over $300{,}000$ parameters) while maintaining convergence performance comparable to FedSGD, making the proposed algorithm well-suited for large-scale problems. This empirical observation further supports the theoretical guarantees in Theorem~\ref{thm:main_2}, where \( m \) is required to grow logarithmically with the dimension \( d \) to retain the \( \mathcal{O}(1/\sqrt{K}) \) convergence rate of FedSGD.

Tables~\ref{tab:fedmpdd_comparison_vrules_2} and \ref{cnn_fedavg_pp_table} demonstrate \texttt{FedMPDD}’s effectiveness in reducing communication cost and its privacy preservation attribute. Our byte budget represents the \emph{total uplink communication} permitted between active clients and the server across all training iterations, unlike per-round limits. We analyze the results from two complementary perspectives, beginning with those reported in Table~\ref{cnn_fedavg_pp_table}.

\smallskip
\textbf{Fixed budget (0.9 GB):}The objective here is to compare methods under a \emph{fixed total uplink communication budget} of $0.9$~GB, evaluating the maximum achievable test accuracy and the corresponding privacy leakage. In Table \ref{cnn_fedavg_pp_table} FedSGD and its Laplace-noised variants 
rapidly exceed the communication budget in the very first iteration, making them impractical under realistic constraints. In contrast, \texttt{FedMPDD} stays well within the budget, achieving competitive accuracy thanks to its efficient projected directional derivative encoding. For example, with $m=600$ (0.2\% of $d$), \texttt{FedMPDD} reaches $40.8\%$ test accuracy, significantly higher than QSGD (12.9\%) and other baselines such as lp-proj (34.7\%)~\cite{lin2022personalized}, Top-k (38.1\%)~\cite{alistarh2018convergence}, and SA-FedLora (35.8\%)~\cite{yang2024spd}. Importantly, although these baselines remain within the budget, they fail to provide consistent privacy guarantees, as their SSIM values (0.74 to 0.91) reveal substantial leakage under gradient inversion attacks. In contrast, \texttt{FedMPDD} achieves both stronger accuracy and substantially lower SSIM (0.14 to 0.22), highlighting its ability to simultaneously reduce communication and preserve privacy, rigorously established in Lemmas~\ref{lem:var_estimator_mpdd_final_adjusted} and~\ref{lem:feature_recovery_bound}, arising from the $(d-m)$-dimensional nullspace of the multi-projected directional derivative."
\smallskip

\textbf{Fixed accuracy (60\% target):} Here, the objective is to compare methods based on the total uplink communication required to reach a fixed target test accuracy of 60\%, while simultaneously evaluating privacy leakage. To achieve the same target accuracy, in Table \ref{cnn_fedavg_pp_table} FedSGD and its noisy variants consume over 470 GB, exceeding the budget by several orders of magnitude and leaking private information (SSIM $>0.8$). Similarly, QSGD requires more than 117 GB while still failing on privacy. lp-proj, Top-k, and SA-FedLora are more communication-efficient, which is their primary goal, requiring only 1.8 to 2.3 GB, but they still exhibit weak privacy protection due to their high SSIM values. By contrast, \texttt{FedMPDD} with $m=600$ requires only 1.3 GB, representing a {more than 356$\times$ reduction} compared to FedSGD, and with $m=2000$, it requires just 3.3 GB, still a {144$\times$ reduction}. Crucially, \texttt{FedMPDD} attains these communication savings while keeping SSIM $<0.22$, ensuring strong and constant privacy level.

Taken together, these results demonstrate that \texttt{FedMPDD} outperforms all baselines across both evaluation criteria: it matches or exceeds their communication efficiency while uniquely combining this with robust privacy protection. Competing methods (lp-proj, Top-k, SA-FedLora) achieve communication reduction but fail on privacy, whereas \texttt{FedMPDD} achieves both simultaneously.

\medskip

Figure~\ref{cifa10_visul_attack} illustrates \texttt{FedMPDD}’s privacy-preserving strength under the GIA \cite{yu2025gi}. The left plot shows SSIM scores over iterations, and the right panel visualizes reconstructed CIFAR-10 samples. Laplace noise with variance $0.1$ (a typical LDP setting) fails to protect data, yielding high SSIM and clear reconstructions, while variance $10$ provides privacy but severely degrades model accuracy. In contrast, \texttt{FedMPDD} with $m=2000$ achieves a comparable privacy level to Laplace($10$) without adding noise, since its protection arises from the $(d-m)$-dimensional nullspace of the projection, rigorously analyzed in Lemmas~\ref{lem:var_estimator_mpdd_final_adjusted} and~\ref{lem:feature_recovery_bound}. At the same time, it reduces per-round communication by more than {150$\times$}, highlighting \texttt{FedMPDD}’s dual benefit of strong privacy and~efficiency.

While increasing  $m$ accelerates convergence, it also incurs higher communication cost and potentially greater privacy leakage (as expected from Lemma~\ref{lem:var_estimator_mpdd_final_adjusted} and \ref{lem:feature_recovery_bound}, increasing $m$ decreases the inherent privacy protection at a rate of $\mathcal{O}(1/m)$, resulting in higher SSIM scores and more successful image reconstructions). However, as illustrated, for instance, in Fig.~\ref{fig:cnn_fmnist_non_iid} in the appendix, smaller values of $m$ can actually achieve comparable or even faster convergence to the target accuracy, while simultaneously offering stronger privacy guarantees as a beneficial side effect. This makes \texttt{FedMPDD} particularly suitable for large-scale problems where both privacy and communication efficiency are critical. This behavior can be intuitively explained by the nullspace effect of the \emph{projected directional derivative} mechanism, which effectively suppresses certain components of noise in the stochastic gradient, thereby stabilizing the optimization. For additional visualizations of another attack model~\cite{zhu2019deep} across different architectures, as well as full training and accuracy curves under various methods, please refer to Appendix~\ref{additional_exp}. 

\section{Conclusion}\label{sec:conclo}
We introduced \texttt{FedMPDD}, a novel FL framework addressing communication efficiency, which also came with some level of privacy protection,  through a gradient encoding and decoding mechanism based on multi-projected directional derivatives. Building upon the single-projection \texttt{FedPDD}, which offered initial communication and privacy benefits but suffered from dimension-dependent convergence, \texttt{FedMPDD} averaged multiple projections to achieve comparable convergence rates to baselines. Our theoretical analysis and empirical evaluations demonstrated \texttt{FedMPDD}'s superior balance of communication cost, performance, and privacy, facilitated by its efficient gradient encoding and decoding. We achieved significant uplink communication reductions compared to baseline methods, including structured, sketched, quantized, and sparsified approaches, while simultaneously ensuring robust and uniform privacy against GIAs, unlike the fluctuating and often weak privacy guarantees of LDP. The tunable parameter $m$ allowed flexible trade-offs. Notably, smaller $m$ values sometimes yielded faster convergence with stronger~privacy.

Several promising avenues remain for extending \texttt{FedMPDD}. First, our current privacy analysis assumes memory-bounded adversaries satisfying $T \times m < d$, a practically relevant model for typical federated training scenarios. Extending guarantees to unbounded-memory adversaries, as commonly modeled in cryptographic~\cite{dwork2006calibrating} and control-theoretic~\cite{teixeira2015secure} frameworks, represents an important theoretical direction. Potential approaches include cryptographic composition with secure aggregation, information-theoretic analysis leveraging gradient evolution during training, or hybrid mechanisms combining projections with differential privacy.
 Second, as noted in Remark~\ref{remark:broader_impact}, client-side computational efficiency can be enhanced through the projected-forward approach, where a fixed mini-batch is sampled once and reused to compute $m$ directional derivatives via forward-mode Jacobian-vector products (JVPs). When $m < \tfrac{hpT}{h + p}$, this approach simultaneously reduces communication overhead, enhances privacy, and lowers computational cost—crucial for resource-constrained federated devices. We plan to implement a fully optimized version leveraging efficient JVPs. Finally, since the projected directional derivative is an unbiased estimator, incorporating momentum and variance reduction techniques offers opportunities to further accelerate convergence while maintaining \texttt{FedMPDD}'s communication and privacy benefits.

\bibliographystyle{ieeetr}%
\bibliography{bib/alias,bib/Reference}

\appendix
\renewcommand{\thesection}{\Alph{section}}
\setcounter{section}{0}
\numberwithin{equation}{subsection}

\subsection{Gradient Inversion Attacks and Deep Leakage from Gradients}\label{DLG:explation}
Gradient inversion attacks (GIAs) are privacy threats in federated learning where an adversary attempts to reconstruct a client's private training data from transmitted gradients~\cite{zhu2019deep, zhao2020idlg, geiping2020inverting, yin2021see, huang2021evaluating}. Since gradients encode information about the input data through the loss function's derivatives, an adversary observing full-dimensional gradients $\vect{g}_i(\vect{x}_k) \in \mathbb{R}^d$ can potentially invert them to recover sensitive information such as images or text. GIAs may operate on a single gradient observation (snapshot) or leverage multiple rounds of observations. These attacks reveal a critical vulnerability in standard FL protocols and motivate the development of privacy-preserving methods.

As a concrete example employed in our empirical evaluations, we describe Deep Leakage from Gradients (DLG)~\cite{zhu2019deep}, an optimization-based attack that reconstructs private training data from a single observed gradient. Given only the observed gradient $\vect{g}_i(\vect{x})$ computed on an \emph{unknown} data pair $(\mathbf{v}, y)$, the attacker reconstructs $(\mathbf{v}, y)$ through the following iterative procedure:

\begin{enumerate}[leftmargin=1.6em]
    \item \emph{Initialization.}  
          The attacker initializes dummy inputs and labels $\mathbf{v}'_1 \sim \mathcal{N}(0, 1)$ and $y'_1 \sim \mathcal{N}(0, 1)$.
          
    \item \emph{Dummy Gradient Computation.}  
          At iteration $i$, the attacker computes the dummy gradient
          \[
              \vect{g}'_i(\vect{x}) = \frac{\partial f\big(F(\mathbf{v}'_i; \vect{x}), y'_i\big)}{\partial \vect{x}},
          \]
          where $F$ is the model architecture and $f$ is the loss function (e.g., cross-entropy).

    \item \emph{Matching Objective.}  
          The discrepancy
          \[
              D_i = \big\| \vect{g}'_i(\vect{x}) - \vect{g}_i(\vect{x}) \big\|_2^2
          \]
          measures how closely the dummy data reproduce the observed gradient.

    \item \emph{Dummy Data Update.} 
          Using gradient-based optimization (commonly Adam or L-BFGS), the attacker updates:
          \[
              \mathbf{v}'_{i+1} = \mathbf{v}'_{i} - \alpha \, \nabla_{\mathbf{v}'} D_i,
              \quad
              y'_{i+1} = y'_{i} - \alpha \, \nabla_{y'} D_i,
          \]
          where $\alpha$ is the learning rate.
\end{enumerate}

After several hundred to a few thousand iterations, the optimized pair $(\mathbf{v}', y')$ can closely approximate the original input. The reconstruction quality depends critically on the information content of the observed gradient: full-dimensional gradients enable highly accurate reconstruction, while obfuscated or compressed gradients reduce the attacker's ability to recover private data. 

DLG and similar GIAs reveal fundamental privacy vulnerabilities in standard FL protocols. This motivates privacy-preserving methods such as differential privacy (adding noise to gradients), secure aggregation (cryptographic masking), or our proposed \texttt{FedMPDD}, which inherently obfuscates gradients through rank-deficient projections that create ambiguity without requiring explicit noise addition or cryptographic overhead.
\subsection{Details and Proofs}\label{sec:proofs}
Below we present omitted proofs and lemmas for the auxiliary and main results presented in the paper. \\

\begin{proof}[Proof of Lemma~\ref{var::reduction:dist_choice}] Recall the variance formula
\begin{align}\label{var_formula}
    \text{Var}[\hat{\vect{g}}(\vect{x})] = \mathbb{E}[\hat{\vect{g}}(\vect{x}) ~\hat{\vect{g}}(\vect{x})^\top] - \mathbb{E}[\hat{\vect{g}}(\vect{x})]\mathbb{E}[\hat{\vect{g}}(\vect{x})]^\top.
\end{align}
We compute $\text{Var}[\hat{\vect{g}}(\vect{x})]$ for the cases where $\vect{u}$ is drawn from either a normal or rademacher distribution. First, we begin with the case where $\vect{u} \sim \mathcal{N}(\mathbf{0}, \mathbf{I}_d)$. Recall, for a random vector $\vect{u} \sim \mathcal{N}(\mathbf{0}, \mathbf{I}_d)$, we have 
\begin{align*}
    \mathbb{E}[\vect{u}] = \vect{0}, ~~\mathbb{E}[\vect{u} \vect{u}^\top] = \mathbf{I}_d. 
\end{align*}
Thus, we have $\mathbb{E}[\hat{\vect{g}}(\vect{x})] = \vect{g}(\vect{x)}$.
Next, compute $\mathbb{E}[\hat{\vect{g}}(\vect{x}) \hat{\vect{g}}(\vect{x})^\top]$ in \eqref{var_formula}
\begin{align}\label{first_term}
    \mathbb{E}[\hat{\vect{g}}(\vect{x}) \hat{\vect{g}}(\vect{x})^\top] = \mathbb{E}[ (\vect{u}^\top \vect{g}(\vect{x}))^2 \vect{u} \vect{u}^\top].
\end{align}
Note that, $\vect{u} \vect{u}^\top = \sum_{m=1}^{d} \sum_{p=1}^{d} u_{m} u_{p} \mathbf{e}_m \mathbf{e}_p^\top 
$ and $(\vect{u}^\top \vect{g}(\vect{x}) )^2 = \left( \sum_{l=1}^d u_{l} g_l(\vect{x}) \right)^2 = \sum_{l=1}^d \sum_{n=1}^d u_{l} u_{n} g_l(\vect{x}) g_n(\vect{x})$, where $\mathbf{e}$ is the basis vector. \\\\
Plugging back into \eqref{first_term}, we have 
\begin{align}
    &\mathbb{E}[\hat{\vect{g}}(\vect{x}) \hat{\vect{g}}(\vect{x})^\top] \nonumber \\
    &= \!\mathbb{E}[ \sum\nolimits_{l=1}^d \!\sum\nolimits_{n=1}^d \!\!\! u_{l} u_{n} g_l(\vect{x}) g_n(\vect{x}) \sum\nolimits_{m=1}^{d} \!\!\sum\nolimits_{p=1}^{d} \!\!\!\!u_{m} u_{p} \mathbf{e}_m \mathbf{e}_p^\top] \nonumber \\
    & =  \!\mathbb{E}[\sum\nolimits_{l=1}^d  \sum\nolimits_{n=1}^d\!  \sum\nolimits_{m=1}^{d} \! \sum\nolimits_{p=1}^{d} u_{l} u_{n} \!g_l(\vect{x}) g_n(\vect{x})   u_{m} u_{p} \mathbf{e}_m \mathbf{e}_p^\top] \nonumber\\ \label{non-zero}
    &= \!\sum\nolimits_{l=1}^d \!\sum\nolimits_{n=1}^d\! \sum\nolimits_{m=1}^{d} \!\sum\nolimits_{p=1}^{d} g_l(\vect{x}) g_n(\vect{x}) \mathbb{E} [u_{l} u_{n} u_{m} u_{p}] \mathbf{e}_m \mathbf{e}_p^\top.
\end{align}
Note that, the last equality comes from the fact that $\vect{u}$ is independent from $\vect{g}(\vect{x})$. Now, there is different cases that $\mathbb{E} [u_{l} u_{n} u_{m} u_{p}]$ is non-zero in \eqref{non-zero}. \\
Case 1 ($l=n$ and $m=p$): In this case, \eqref{non-zero} simplifies to the following 
\begin{align*}
    &\mathbb{E}[\hat{\vect{g}}(\vect{x}) \hat{\vect{g}}(\vect{x})^\top] \!=\!\!  \sum_{l=1}^d  \sum_{m=1}^{d} g_l^{2}(\vect{x}) \mathbb{E} [u_{l}^2] \mathbb{E} [u_{m}^2] \mathbf{e}_m \mathbf{e}_m^\top \!\!= \!\! \|\vect{g}(\vect{x)} \|^2 \mathbf{I}_d.
\end{align*}
Case 2 ($l=m$ and $n=p$): In this case, \eqref{non-zero} simplifies to the following 
\begin{align*}
    \mathbb{E}[\hat{\vect{g}}(\vect{x}) \hat{\vect{g}}(\vect{x})^\top] &=  \sum_{l=1}^d \sum_{n=1}^d g_l(\vect{x}) g_n(\vect{x}) \mathbb{E} [u_{l}^2] \mathbb{E}[u_{n}^2] \mathbf{e}_l \mathbf{e}_n^\top \nonumber \\
    &=  \vect{g}(\vect{x}) (\vect{g}(\vect{x}))^\top
\end{align*}

Case 3 ($l=p$ and $m=n$): Similar to case 2, \eqref{non-zero} simplifies to the following 
\begin{align*}
    \mathbb{E}[\hat{\vect{g}}(\vect{x}) \hat{\vect{g}}(\vect{x})^\top] =  \vect{g}(\vect{x}) (\vect{g}(\vect{x}))^\top
\end{align*}

Case 4 ($l=n=m=p$): In this case, \eqref{non-zero} simplifies to the following 
\begin{align*}
     &\mathbb{E}[\hat{\vect{g}}(\vect{x}) \hat{\vect{g}}(\vect{x})^\top] = \sum\nolimits_{l=1}^d  g_l^2(\vect{x}) \mathbb{E} [u_{l}^4] \mathbf{e}_l \mathbf{e}_l^\top = 3 \|\vect{g}(\vect{x})\|^2 \mathbf{I}_d
\end{align*}

Then, \eqref{non-zero} can be simplified as follows 
\begin{align*}
    \mathbb{E}[\hat{\vect{g}}(\vect{x}) \hat{\vect{g}}(\vect{x})^\top] = 2 ~\vect{g}(\vect{x}) (\vect{g}(\vect{x}))^\top + 4  \|\vect{g}(\vect{x})\|^2 \mathbf{I}_d.
\end{align*}

As a result, \eqref{var_formula} can be written as follows for the case where $\vect{u}$ drawn from a normal distribution with zero mean and unit variance.
\begin{align}\label{final_normal}
    \text{Var}_{\vect{u} \sim \mathcal{N}(\mathbf{0}, \mathbf{I}_d)}[\hat{\vect{g}}(\vect{x})] =  ~\vect{g}(\vect{x}) (\vect{g}(\vect{x}))^\top + 4  \|\vect{g}(\vect{x})\|^2 \mathbf{I}_d.
\end{align}

Now, we compute the variance \( \text{Var}[\hat{\vect{g}}(\vect{x})] \) in the case where \( \vect{u} \sim \text{Rademacher}^d \). 
Recall that for a random vector \( \vect{u} \) with i.i.d.\ Rademacher entries, we have 
$
\mathbb{E}[\vect{u}] = \vect{0} \quad \text{and} \quad \mathbb{E}[\vect{u} \vect{u}^\top] = \mathbf{I}_d.
$
Therefore, we have $
\mathbb{E}[\hat{\vect{g}}(\vect{x})] = \vect{g}(\vect{x}).$

Similar to the proof for the normal distribution, cases $1, 2$, and $3$ are identical. However in the fourth case, where $l=n=m=p$, since the fourth moment of a Rademacher distribution is $1$, we get $\mathbb{E}[\hat{\vect{g}}(\vect{x}) \hat{\vect{g}}(\vect{x})^\top] =   \|\vect{g}(\vect{x})\|^2 \mathbf{I}_d$. Then, \eqref{non-zero} can be simplified as follows 
\begin{align*}
    \mathbb{E}[\hat{\vect{g}}(\vect{x}) \hat{\vect{g}}(\vect{x})^\top] = 2 ~\vect{g}(\vect{x}) (\vect{g}(\vect{x}))^\top + 2 \|\vect{g}(\vect{x)}\|^2 \mathbf{I}_d.
\end{align*}

Thus, we get
\begin{align}\label{final_radem}
    \text{Var}_{\vect{u} \sim \text{Rademacher}^d}[\hat{\vect{g}}(\vect{x})] =  \vect{g}(\vect{x}) (\vect{g}(\vect{x}))^{\top} + 2  \|\vect{g}(\vect{x})\|^2 \mathbf{I}_d.
\end{align}

From \eqref{final_normal} and \eqref{final_radem}, we have the following
\begin{align}
    \text{Var}_{\vect{u} \sim \mathcal{N}(\mathbf{0}, \mathbf{I}_d)}[\hat{\vect{g}}(\vect{x})] & - \text{Var}_{\vect{u} \sim \text{Rademacher}^d}[\hat{\vect{g}}(\vect{x})]= 2  \|\vect{g}(\vect{x})\|^2 \mathbf{I}_d
\end{align}
which concludes the proof.
\end{proof}

\begin{proof}[Proof of Lemma \ref{lem:var_estimator_mpdd_final_adjusted}]
Note that for a single direction $\vect{u}_{k,i}$ ($m=1$),

\begin{align}
\bigl\|\hat{\vect g}_i(\vect x_k)\bigr\|^2
&= \vect u_{k,i}^\top
   \bigl(\vect u_{k,i}^\top \vect g_i(\vect x_k)\bigr)
   \bigl(\vect u_{k,i}^\top \vect g_i(\vect x_k)\bigr)
   \vect u_{k,i} \nonumber\\
&= (\vect u_{k,i}^\top \vect u_{k,i})
   \bigl(\vect g_i(\vect x_k)^\top
         \vect u_{k,i}\vect u_{k,i}^\top
         \vect g_i(\vect x_k)\bigr).
\end{align}
Since for Rademacher $\vect u_{k,i}$, we have $\vect u_{k,i}^\top \vect u_{k,i}=d$, it follows that
\[
\mathbb{E}_{\vect u}\!\left[\|\hat{\vect g}_i(\vect x_k)\|^2\right]
\!\!= \!d\, \vect g_i(\vect x_k)^\top \,\!\!\mathbb{E}_{\vect u}[\vect u_{k,i}\vect u_{k,i}^\top]\, \vect g_i(\vect x_k)
\!= \!\!d\, \|\vect g_i(\vect x_k)\|^2.
\]

Therefore, since the estimator satisfies
\[
\mathbb{E}_{\vect u}\!\left[\hat{\vect g}_i(\vect x_k)\right] = \vect g_i(\vect x_k),
\]

we obtain for one direction
\begin{align}
\mathbb{E}_{\vect u}\!\left[\|\hat{\vect g}_i(\vect x_k) - \vect g_i(\vect x_k)\|^2\right]
&= \mathbb{E}_{\vect u}\!\left[\|\hat{\vect g}_i(\vect x_k)\|^2 - \|\vect g_i(\vect x_k)\|^2\right] \nonumber\\
&= (d-1)\|\vect g_i(\vect x_k)\|^2.
\end{align}

Now extend to $m$ directions. The multi-direction estimator is
\[
\hat{\vect g}_i(\vect x_k) = \frac{1}{m}\sum\nolimits_{j=1}^m \vect{u}_{k,i}^{(j)}(\vect{u}_{k,i}^{(j)})^\top \vect g_i(\vect x_k).
\]
Because the $u_{k,i}^{(j)}$ are independent and identically distributed, the cross terms vanish in expectation, and all diagonal terms are the same. Thus,
\[
\mathbb{E}\!\left[\|\hat{\vect g}_i(\vect x_k) - \vect g_i(\vect x_k)\|^2\right]
= \frac{1}{m}\,(d-1)\,\|\vect g_i(\vect x_k)\|^2.
\]

Dividing both sides by $\|\vect g_i(\vect x_k)\|^2$ yields in \eqref{eq:relative_error_mpdd_final_adjusted}.
\end{proof}

\begin{proof}[Proof of Lemma \ref{lem:feature_recovery_bound}]
Consider the reconstruction loss
\[
\begin{aligned}
\mathcal L(\hat v)
&= \left\|\tfrac{1}{m} \mathbf{U}_{k,i} \mathbf{U}_{k,i}^\top \vect{g}_i(v,c;\vect{x}_k)-\vect{g}_i(\hat v,c;\vect{x}_k)\right\| \\
&= \Big\|\big(\vect{g}_i(v,c;\vect{x}_k)-\vect{g}_i(\hat v,c;\vect{x}_k)\big) -\nonumber\\
&~~~~\big(\vect{g}_i(v,c;\vect{x}_k)-\tfrac{1}{m}\mathbf{U}_{k,i} \mathbf{U}_{k,i}^\top \vect{g}_i(v,c;\vect{x}_k)\big)\Big\| \\
&\ge \Big|\;\|\vect{g}_i(v,c;\vect{x}_k)-\tfrac{1}{m}\mathbf{U}_{k,i} \mathbf{U}_{k,i}^\top \vect{g}_i(v,c;\vect{x}_k)\|
- \nonumber\\
&~~~~\|\vect{g}_i(v,c;\vect{x}_k)-\vect{g}_i(\hat v,c;\vect{x}_k)\|\;\Big| \\
&\ge \|(I-\tfrac{1}{m} \mathbf{U}_{k,i}\mathbf{U}_{k,i}^\top)\vect{g}_i(v,c;\vect{x}_k)\|
- \nonumber\\
&~~~~\|\vect{g}_i(v,c;\vect{x}_k)-\vect{g}_i(\hat v,c;\vect{x}_k)\|.
\end{aligned}
\]
Thus,
$
\|\vect{g}_i(v,c;\vect{x}_k)-\vect{g}_i(\hat v,c;\vect{x}_k)\|\;\ge\;\operatorname{proj}_g(v,c;\vect{x}_k,\mathbf{U}_{k,i})-\mathcal L(\hat v),
$
where $
\operatorname{proj}_g(v,c;\vect{x}_k,\mathbf{U}_{k,i}) := \|(I- \tfrac{1}{m}\mathbf{U}_{k,i}\mathbf{U}_{k,i}^\top)\,\vect{g}_i(v,c;\vect{x}_k)\|$ 
denotes the projection-induced gradient error.  

Assuming that for fixed $\vect{x}_k$ and known $c$, the map $v \mapsto \vect{g}_i(v,c;\vect{x}_k)$ is $L_v(\vect{x}_k)$-Lipschitz (as is satisfied by standard models), we obtain
\[
\|v-\hat v\|\;\ge\;\frac{\operatorname{proj}_g(v,c;\vect{x}_k,\mathbf{U}_{k,i})-\mathcal L(\hat v)}{L_v(\vect{x}_k)}.
\]
With the convention that the right-hand side is interpreted as $\max\{0,\cdot\}$, in particular for an attack-optimal recovery $\hat v^\star\in\arg\min_{\hat v}\mathcal L(\hat v)$,
\begin{align}
\|v&-\hat v^\star\|
\ge \frac{\operatorname{proj}_g(v,c;\vect{x}_k,\mathbf{U}_{k,i})-\mathcal L(\hat v^\star)}{L_v(\vect{x}_k)}, \nonumber\\
&\qquad \text{and if }\mathcal L(\hat v^\star)=0,\;\;
\|v-\hat v^\star\|
\ge \frac{\operatorname{proj}_g(v,c;\vect{x}_k,\mathbf{U}_{k,i})}{L_v(\vect{x}_k)}.
\end{align}

Squaring the above bound, taking expectation with respect to $\mathbf{U}_{k,i}$, and invoking Lemma~\ref{lem:var_estimator_mpdd_final_adjusted}, we arrive at
\begin{align}
\mathbb{E}\!\left[\|v-\hat v^\star\|^2\right]
&\ge \frac{1}{L_v(\vect{x}_k)^2}\,
\mathbb{E}\!\left[\operatorname{proj}_g(v,c;\vect{x}_k,\mathbf{U}_{k,i})^2\right] \nonumber\\
&= \frac{d-1}{m\,L_v(\vect{x}_k)^2}\,
\|\vect{g}_i(v,c;\vect{x}_k)\|^2.
\end{align}

Here, we used the fact that from Lemma \ref{lem:var_estimator_mpdd_final_adjusted}, i.e., 
\[
\mathbb{E}\!\left[\operatorname{proj}_g(v,c;\vect{x}_k,\mathbf{U}_{k,i})^2\right]
= \frac{d-1}{m}\,\|\vect{g}_i(v,c;\vect{x}_k)\|^2.
\]

\end{proof}

\begin{proof}[Proof of Theorem \ref{thm:main}]
 Since $f$ is $L-smooth$, we can write
\begin{align}\label{eq::1}
     \mathbb{E}[f(\vect{x}_{k+1})|\vect{x}_{k}] \leq f(\vect{x}_{k})+\big \langle\nabla f(\vect{x}_{k}), \mathbb{E}[\vect{x}_{k+1} - \vect{x}_{k}|\vect{x}_{k}] \big \rangle \nonumber\\+\frac{L}{2}\mathbb{E}\big[\|\vect{x}_{k+1} - \vect{x}_{x}\|^2\big| \vect{x}_{k}\big],
\end{align}
from line $14$ of \texttt{FedPDD} algorithm into \eqref{eq::1}, we have 
\begin{align}\label{eq::g_hat}
     &\mathbb{E}[f(\vect{x}_{k+1})|\vect{x}_{k}] \leq f(\vect{x}_{k}) -\eta ~\big \langle\nabla f(\vect{x}_{k}), \mathbb{E}[\hat{\vect{g}}(\vect{x}_k)|\vect{x}_{k}] \big \rangle \nonumber \\
     &+\frac{L\eta^2}{2}\mathbb{E}\big[\|\hat{\vect{g}}(\vect{x}_k) \|^2\big| \vect{x}_{k}\big],\nonumber\\
     & = f(\vect{x}_{k}) -\eta ~\|\nabla f(\vect{x}_{k})\|^2 +\frac{L\eta^2}{2}\mathbb{E}\big[\|\hat{\vect{g}}(\vect{x}_k) \|^2\big| \vect{x}_{k}\big],\nonumber\\
     & = f(\vect{x}_{k}) -\eta ~\|\nabla f(\vect{x}_{k})\|^2 +\frac{L\eta^2}{2} \big( \mathbb{E}\big[\|\hat{\vect{g}}(\vect{x}_k)  - \nabla f(\vect{x}_k)\|^2\big| \vect{x}_{k}\big] \nonumber\\
     &+ \|\nabla f(\vect{x}_k)\|^2 \big),\nonumber\\
      & = f(\vect{x}_{k}) -(\eta - \frac{L\eta^2}{2}) ~\|\nabla f(\vect{x}_{k})\|^2 \nonumber\\
      &+\frac{L\eta^2}{2}\mathbb{E}\big[\|\hat{\vect{g}}(\vect{x}_k) - \nabla f(\vect{x}_k)\|^2\big| \vect{x}_{k}\big],
\end{align}
where the first equality comes from the unbiasedness of $\hat{\vect{g}}(\vect{x}_k)$, i.e., $\mathbb{E}[\hat{\vect{g}}(\vect{x}_k)|\vect{x}_{k}] = \nabla f(\vect{x}_k)$, the second equality comes from bias–variance decomposition of $\mathbb{E}[\|\hat{\vect{g}}(\vect{x}_k)\|^2\big|\vect{x}_k]$. Provided that $\eta < \frac{1}{L}$, and taking total expectation on both sides of \eqref{eq::g_hat} we have
\begin{align}\label{eq::g_hat_bound}
     \mathbb{E}[f(\vect{x}_{k+1})] & \leq \mathbb{E}[f(\vect{x}_{k})] -\frac{\eta}{2} ~\mathbb{E}[\|\nabla f(\vect{x}_{k})\|^2] \nonumber\\
     &+\frac{L\eta^2}{2}\mathbb{E}\big[\|\hat{\vect{g}}(\vect{x}_k) - \nabla f(\vect{x}_k)\|^2\big],
\end{align}
First we simplify the second summand in the upper bound of \eqref{eq::g_hat_bound} by adding and subtracting $\frac{1}{\beta N }\sum_{i \in \mathcal{A}_k} \vect{g}_i(\vect{x}_k)$, which leads to
\begin{align}\label{eq:add_subtract}
      \mathbb{E}\big[\|&\hat{\vect{g}}(\vect{x}_k) - \nabla f(\vect{x}_k)\|^2\big] = \mathbb{E}\big[\| \frac{1}{\beta N } \sum\nolimits_{i \in \mathcal{A}_k} \vect{g}_i(\vect{x}_k) - \nabla f(\vect{x}_k) \nonumber \\
      &+ \hat{\vect{g}}(\vect{x}_k) - \frac{1}{\beta N } \sum\nolimits_{i \in \mathcal{A}_k} \vect{g}_i(\vect{x}_k)  \|^2\big] \nonumber\\
      &\leq 2\mathbb{E}\big[\| \frac{1}{\beta N } \sum\nolimits_{i \in \mathcal{A}_k} \vect{g}_i(\vect{x}_k) - \nabla f(\vect{x}_k)\|^2\big]
      + 2\mathbb{E}[\|\hat{\vect{g}}(\vect{x}_k) \nonumber\\
      &- \frac{1}{\beta N } \sum\nolimits_{i \in \mathcal{A}_k} \vect{g}_i(\vect{x}_k)  \|^2\big] \nonumber\\
      &= 2\mathbb{E}\big[\| \frac{1}{\beta N } \sum\nolimits_{i \in \mathcal{A}_k} \vect{g}_i(\vect{x}_k) - \nabla f(\vect{x}_k)\|^2\big]
      \nonumber \\
      &+ 2\mathbb{E}[\| \frac{1}{\beta N} \sum\nolimits_{i \in \mathcal{A}_k} s_i^k \vect{u}_{k,i} - \frac{1}{\beta N } \sum\nolimits_{i \in \mathcal{A}_k} \vect{g}_i(\vect{x}_k)  \|^2\big] \nonumber\\
      &\leq 2\mathbb{E}\big[\| \frac{1}{\beta N } \sum\nolimits_{i \in \mathcal{A}_k} \vect{g}_i(\vect{x}_k) - \nabla f(\vect{x}_k)\|^2\big]
     \nonumber\\
     &+ \frac{2}{\beta N} \sum\nolimits_{i \in \mathcal{A}_k} \mathbb{E}[\| s_i^k \vect{u}_{k,i} - \vect{g}_i(\vect{x}_k)  \|^2\big],
\end{align}
where the first inequality follows from $||a+b||^2 \leq 2 \|a\|^2 + 2\|b\|^2$, and the second inequality follows from Jensen’s inequality. 

Next, we note that using~\cite[Lemma 4]{t2020personalized} we can write
\begin{align}
\mathbb{E}_{\mathcal{A}_k}& \left\| \frac{1}{\beta N} \sum\nolimits_{i \in \mathcal{A}_k} \nabla f_i(\vect{x}_k) - \nabla f(\vect{x}_k) \right\|_2^2
\leq \nonumber \\
&\frac{1/\beta - 1}{N-1} \cdot \frac{1}{N} \sum\nolimits_{i=1}^N \left\| \nabla f_i(\vect{x}_k) - \nabla f(\vect{x}_k) \right\|_2^2,
\end{align}
which along with~\eqref{eq:add_subtract} leads to 
\begin{align}\label{eq:bound_g_hat-g}
    \mathbb{E}\big[\|&\hat{\vect{g}}(\vect{x}_k) - \nabla f(\vect{x}_k)\|^2\big] \leq  \frac{2(1/\beta - 1)}{N(N-1)} \sum\nolimits_{i=1}^{N} \mathbb{E} [\| \nabla f_i(\vect{x}_{k}) - \nonumber\\
    &\nabla f(\vect{x}_k)\|^2]) + \frac{2}{\beta N} \sum\nolimits_{i \in \mathcal{A}_k} \mathbb{E}[\| s_i^k \vect{u}_{k,i} - \vect{g}_i(\vect{x}_k)  \|^2\big], \nonumber\\
      &= \frac{2(1/\beta - 1)}{N(N-1)} \sum\nolimits_{i=1}^{N} \mathbb{E} [\| \nabla f_i(\vect{x}_{k}) - \nabla f(\vect{x}_k)\|^2]) \nonumber\\
      &+ \frac{2}{\beta N} \sum\nolimits_{i \in \mathcal{A}_k} \mathbb{E}[\| \hat{\vect{g}}_i(\vect{x}_k) - \vect{g}_i(\vect{x}_k)  \|^2\big], 
\end{align}
where the equality comes from the fact that $s_i^k \vect{u}_{k,i} = \hat{\vect{g}}_i(\vect{x}_k)$. Now, using the result from Lemma \ref{lem:var_estimator_mpdd_final_adjusted},  \eqref{eq:bound_g_hat-g} leads to 

\begin{align}\label{fin_bound_second_term}
    \mathbb{E}\big[\|&\hat{\vect{g}}(\vect{x}_k) - \nabla f(\vect{x}_k)\|^2\big] \leq \frac{2(1/\beta - 1)}{N(N-1)} \sum\nolimits_{i=1}^{N} \mathbb{E} [\| \nabla f_i(\vect{x}_{k}) \nonumber\\
    &- \nabla f(\vect{x}_k)\|^2]) + \frac{2(d-1)}{\beta N} \sum\nolimits_{i \in \mathcal{A}_k} \mathbb{E}[\|\vect{g}_i(\vect{x}_k)  \|^2\big]. 
\end{align}
Subsequently, given \eqref{fin_bound_second_term},  we obtain from \eqref{eq::g_hat_bound}:
\begin{align}\label{eq::g_hat_bound_final}
     &\mathbb{E}[f(\vect{x}_{k+1})]  \leq \mathbb{E}[f(\vect{x}_{k})] -\frac{\eta}{2} ~\mathbb{E}[\|\nabla f(\vect{x}_{k})\|^2] \nonumber \\
     & +\frac{L\eta^2}{2}\bigg( \frac{2(1/\beta - 1)}{N(N-1)} \sum\nolimits_{i=1}^{N} \mathbb{E} [\| \nabla f_i(\vect{x}_{k}) - \nabla f(\vect{x}_k)\|^2])  \nonumber\\
     &+\frac{2(d-1)}{\beta N} \sum\nolimits_{i \in \mathcal{A}_k} \mathbb{E}[\|\vect{g}_i(\vect{x}_k)  \|^2\big] \bigg).
\end{align}
Invoking Assumption~\ref{assump:smooth}, which bounds the variance of local gradients relative to the global gradient in \eqref{eq::g_hat_bound_final}, we get
\begin{align}\label{eq::g_hat_bound_final_assump}
     &\mathbb{E}[f(\vect{x}_{k+1})]  \leq \mathbb{E}[f(\vect{x}_{k})] -\frac{\eta}{2} ~\mathbb{E}[\|\nabla f(\vect{x}_{k})\|^2]+ \nonumber \\
     & \frac{L\eta^2}{2}\bigg( \frac{2(1/\beta - 1)}{N-1} \sigma^2  + \frac{2(d-1)}{\beta N} \sum\nolimits_{i \in \mathcal{A}_k} \mathbb{E}[\|\vect{g}_i(\vect{x}_k)  \|^2\big] \bigg).\nonumber \\
\end{align}
Rearranging and summing $k$ from $0$ to $K - 1$ in \eqref{eq::g_hat_bound_final_assump}, we have 
\begin{align}\label{final_boundd}
    &\frac{1}{K} \sum\nolimits_{k=0}^{K-1} \mathbb{E}[\|\nabla f(\vect{x}_{k})\|^2] \leq \frac{2}{ K \eta}(\mathbb{E}[f(\vect{x}_{0})] - \mathbb{E}[f(\vect{x}_{K})])  + \nonumber\\
    &\frac{2 L \eta  (1/\beta - 1)}{K(N-1)} \sigma^2 + \frac{2 L\eta (d-1)}{K \beta N } \sum\nolimits_{k=0}^{K-1} \sum\nolimits_{i \in \mathcal{A}_k} \mathbb{E}\big[\|\vect{g}_i(\vect{x}_k)\|^2] 
    \nonumber \\
    &\leq \frac{2}{ K \eta}(\mathbb{E}[f(\vect{x}_{0})] - f^\star)  + \frac{2 L \eta  (1/\beta - 1)}{K(N-1)} \sigma^2 \nonumber\\
    &+ \frac{2 L\eta (d-1)}{K \beta N } \sum\nolimits_{k=0}^{K-1} \sum\nolimits_{i \in \mathcal{A}_k} \mathbb{E}\big[\|\vect{g}_i(\vect{x}_k)\|^2].
\end{align}
We simplify the upper bound in~\eqref{final_boundd} by invoking Assumption~\ref{assump:smooth} on the bounded variance of the stochastic gradient, we get
\begin{align}\label{result:thm1_neww}
    \frac{1}{K} \sum\nolimits_{k=0}^{K-1} \mathbb{E}[\|\nabla &f(\vect{x}_{k})\|^2]  \leq \frac{2}{ K \eta} (f(\vect{x}_{0}) - f^\star) \nonumber\\
    &+ \frac{2 L \eta  (1/\beta - 1)}{K(N-1)} \sigma^2   
    +2 L \eta d G^2 .
\end{align}
If we set $\eta = \frac{1}{L\sqrt{K}}$ in \eqref{result:thm1_neww}, we have a convergence rate of $\mathcal{O}(d/\sqrt{K})$ to a stationary point of $f(\vect{x})$.
\end{proof}


\begin{proof}[Proof of Theorem \ref{thm:main_2}]
For notational convenience, we stack the \(m\) random direction vectors into a single matrix
\(
\mathbf{U}_{k,i} = \bigl[\mathbf{u}_{k,i}^{(1)} \;\mathbf{u}_{k,i}^{(2)} \;\dots \;\mathbf{u}_{k,i}^{(m)}\bigr] \in \mathbb{R}^{d \times m},
\) which simplifies the analysis that follows. With this compact form, the gradient estimator in \texttt{FedMPDD} algorithm \( \hat{\mathbf{g}}_i(\mathbf{x}_k) = \frac{1}{m} \mathbf{U}_{k,i} \mathbf{U}_{k,i}^{\!\top} \mathbf{g}_i(\mathbf{x}_k) \) remains unbiased, i.e.,
$
\mathbb{E}\left[\hat{\mathbf{g}}_i(\mathbf{x}_k)\right] 
= \frac{1}{m} \mathbb{E}\left[\mathbf{U}_{k,i} \mathbf{U}_{k,i}^{\!\top}\right] \mathbf{g}_i(\mathbf{x}_k) 
= \mathbf{g}_i(\mathbf{x}_k),
$
most of the derivation steps leading up to Equation~\eqref{eq:bound_g_hat-g} remain unchanged. Therefore, we omit those details for brevity. From~\eqref{eq:bound_g_hat-g} in the proof of Theorem~\ref{thm:main}, we have
\begin{align}\label{eq:bound_g_hat-g_alg_2}
   & \mathbb{E}\big[\|\hat{\vect{g}}(\vect{x}_k) - \nabla f(\vect{x}_k)\|^2\big] \leq \frac{2(1/\beta - 1)}{N(N-1)} \sum\nolimits_{i=1}^{N} \mathbb{E} [\| \nabla f_i(\vect{x}_{k}) - \nonumber \\
   &\nabla f(\vect{x}_k)\|^2] + \frac{2}{\beta N} \sum\nolimits_{i \in \mathcal{A}_k} \mathbb{E}[\| \hat{\vect{g}}_i(\vect{x}_k) - \vect{g}_i(\vect{x}_k)  \|^2\big], \nonumber\\
    &= \frac{2(1/\beta - 1)}{N(N-1)} \sum\nolimits_{i=1}^{N} \mathbb{E} [\| \nabla f_i(\vect{x}_{k}) - \nabla f(\vect{x}_k)\|^2] \nonumber \\
    &+ \frac{2}{\beta N} \sum\nolimits_{i \in \mathcal{A}_k} \mathbb{E}[\| \hat{\vect{g}}_i(\vect{x}_k)\|^2] - \mathbb{E}[\|\vect{g}_i(\vect{x}_k)  \|^2], 
\end{align}
the last equality holds because the estimator is unbiased and the Rademacher directions are sampled independently of all other sources of randomness. Where \( \hat{\mathbf{g}}(\mathbf{x}_k) = \frac{1}{m \beta N}  \sum_{i\in\mathcal{A}_k} \mathbf{U}_{k,i} \mathbf{U}_{k,i}^{\!\top}\mathbf{g}_i(\mathbf{x}_k) \), with \( \beta N = |\mathcal{A}_k| \) denoting the number of participating clients at round \( k \), and \( \beta \in (0, 1] \) representing the client sampling fraction. Using JL Lemma \ref{lem:opJL} results, for $m=\mathcal{O} \bigl(\tfrac{\ln(d/\delta)}{\varepsilon^{2}}\bigr)$ into \eqref{eq:bound_g_hat-g_alg_2} with probability at least $1-\delta$, we have 
\begin{align}\label{eq:bound_g_hat-g_alg_2_new}
    \mathbb{E}\big[\|\hat{\vect{g}}(\vect{x}_k) &- \nabla f(\vect{x}_k)\|^2\big] 
    \leq \frac{2(1/\beta - 1)}{N(N-1)} \sum\nolimits_{i=1}^{N} \mathbb{E} [\| \nabla f_i(\vect{x}_{k}) -\nonumber \\ & \nabla f(\vect{x}_k)\|^2] 
    + \frac{\epsilon(4 + 2\epsilon)}{\beta N} \sum\nolimits_{i \in \mathcal{A}_k} \mathbb{E}[\|\vect{g}_i(\vect{x}_k)  \|^2]. 
\end{align}
Now, similar to the proof steps of Theorem \ref{thm:main} following the steps after equation \eqref{eq:bound_g_hat-g} and invoking Assumption \ref{assump:smooth}, we obtain the following upper bound
\begin{align}\label{result:thm1}
    \frac{1}{K} \sum\nolimits_{k=0}^{K-1} \mathbb{E}[\|\nabla &f(\vect{x}_{k})\|^2]  \leq \frac{2}{ K \eta} (f(\vect{x}_{0}) - f^\star) \nonumber\\
    &+ \frac{2 L \eta  (1/\beta - 1)}{K(N-1)} \sigma^2   
    + \epsilon (4+ 2\epsilon) L \eta  G^2,
\end{align}
where $0<\epsilon <1$ is the distortion parameter. If we set $\eta = \frac{1}{L\sqrt{K}}$  in \eqref{result:thm1}, we have a convergence rate of $\mathcal{O}(1/\sqrt{K})$ to a stationary point of $f(\vect{x})$.
    
\end{proof}

\begin{proof}[Proof of Lemma \ref{thm:worst_case_privacy}]
In the worst case where $g_i(x^k) = g$ is constant, each round $k$ provides $m$ linear equations of the form $\langle g, u_{i}^{(j)} \rangle = s_i^{(j)}$ for $j = 1, \ldots, m$. After $T$ rounds, the adversary has $T \times m$ linear constraints on the $d$-dimensional vector $g$. The system is underdetermined when $T \times m < d$, ensuring that multiple solutions exist in the $(d - T \times m)$-dimensional nullspace, preventing unique gradient recovery.
\end{proof}

\subsection{Hyperparameters}\label{hyper_parameters}
This appendix summarizes the hyperparameters used for each baseline method and our proposed algorithm. To ensure reproducibility, we selected five fixed random seeds as $[17, 123, 777, 2023, 424242]$. Additionally, we use a separate seed, $2024$, to control client data partitioning.

\begin{itemize}
   \item  \textbf{Logistic model on the MNIST dataset.} See Table~\ref{tab:hyperparams_logisitc_mnist}. The best stepsize for each algorithm is selected from the set $\{0.1, 0.01, 0.001\}$.
   \item \textbf{CNN model from \cite{lin2022personalized} on the MNIST and FASHIONMNIST datasets.} See Table~\ref{tab:hyperparams_cnn_70}. The best stepsize for each algorithm is selected from the set $\{0.1, 0.01, 0.001\}$.
   \item \textbf{CNN model from \cite{mcmahan2017communication} on CIFAR10 dataset.} See Table~\ref{tab:hyperparams_cnn_fedavg}. The best stepsize for each algorithm is selected from the set $\{0.01, 0.005, 0.0001\}$.
   \item \textbf{LeNet model on the MNIST and FASHIONMNIST datasets.} See Table~\ref{tab:hyperparams_lenet}. The best stepsize for each algorithm is selected from the set $\{0.1, 0.01, 0.001\}$.
\end{itemize}

For the DLG attack hyper-parameters, we followed the procedure in \cite{zhu2019deep}. To improve stability, we used the Adam optimizer for both the logistic-regression model and the CNN from \cite{mcmahan2017communication}, and the L-BFGS optimizer (history size 100, max 20 iterations) for the LeNet model and the CNN from \cite{lin2022personalized}.    






\begin{table}[t]
\centering
\scriptsize
\caption{\small{Hyperparameters used for all methods on the logistic regression model.}}
\label{tab:hyperparams_logisitc_mnist}
\begin{tabular}{lcccc}
\toprule
\textbf{Method} & \textbf{batch} & \textbf{lr} & \textbf{opt.} & \textbf{client participation (\%)}\\
\midrule
FedSGD                 & 1 & 0.01     & sgd  & 10 \\
QSGD             & 1 & 0.01  & sgd & 10 \\
\textbf{FedMPDD}          & 1 & 0.01  & sgd & 10 \\
FedSGD + Gaussian          & 1 & 0.01  & sgd & 10 \\
FedSGD + Laplace         & 1 & 0.01  & sgd & 10 \\
\bottomrule
\end{tabular}
\end{table}

\begin{table}[t]
\centering
\scriptsize
\caption{Hyperparameters used for all methods on the CNN model from \cite{lin2022personalized}.}
\label{tab:hyperparams_cnn_70}
\begin{tabular}{lcccc}
\toprule
\textbf{Method} & \textbf{batch} & \textbf{lr} & \textbf{opt.} & \textbf{client participation (\%)}\\
\midrule
FedSGD                 & 64 & 0.1     & sgd  & 50 \\
QSGD             & 64 & 0.1  & sgd & 50 \\
\textbf{FedMPDD}          & 64 & 0.1  & sgd & 50 \\
FedSGD + Laplace         & 64 & 0.1  & sgd & 50 \\
\bottomrule
\end{tabular}
\end{table}


\begin{table}[t]
\centering
\scriptsize
\caption{Hyperparameters used for all methods on the CNN model from \cite{mcmahan2017communication}.}
\label{tab:hyperparams_cnn_fedavg}
\begin{tabular}{lcccc}
\toprule
\textbf{Method} & \textbf{batch} & \textbf{lr} & \textbf{opt.} & \textbf{client participation (\%)}\\
\midrule
FedSGD                 & 64 & 0.005     & sgd  & 100 \\
QSGD                   & 64 & 0.005     & sgd  & 100 \\
\textbf{FedMPDD}       & 64 & 0.005     & sgd  & 100 \\
FedSGD + Laplace       & 64 & 0.005     & sgd  & 100 \\
Top-k                  & 64 & 0.005     & sgd  & 100 \\
lp-proj                & 64 & 0.005     & sgd  & 100 \\
SA-FedLora             & 64 & 0.005     & sgd  & 100 \\
\bottomrule
\end{tabular}
\end{table}

\begin{table}[t]
\centering
\scriptsize
\caption{Hyperparameters used for all methods on the LeNet model.}
\label{tab:hyperparams_lenet}
\begin{tabular}{lcccc}
\toprule
\textbf{Method} & \textbf{batch} & \textbf{lr} & \textbf{opt.} & \textbf{client participation (\%)}\\
\midrule
FedSGD                 & 1 & 0.1     & sgd  & 50 \\
QSGD                   & 1 & 0.1     & sgd  & 50 \\
\textbf{FedMPDD}       & 1 & 0.1     & sgd  & 50 \\
FedSGD + Laplace       & 1 & 0.1     & sgd  & 50 \\
Top-k           & 1 & 0.1     & sgd  & 50 \\
lp-proj                & 1 & 0.1     & sgd  & 50 \\
\bottomrule
\end{tabular}
\end{table}

\subsubsection{License Information for Datasets}
\textbf{CIFAR10.} The original CIFAR10 dataset is available under the MIT license.

\textbf{MNIST.} The original MNIST dataset is available under the CC BY-SA 3.0 license.

\textbf{FASHIONMNIST.} The original FASHIONMNIST dataset is available under the MIT license.


\subsection{Additional Experimental Evaluation Results}\label{additional_exp}
This appendix presents additional experimental results on both i.i.d. and non-i.i.d.
data distributions, evaluating the proposed method in terms of communication cost
reduction and privacy preservation across various datasets and models with varying
dimensionalities.

\begin{figure*}[t]
\centering
\includegraphics[scale=0.3]{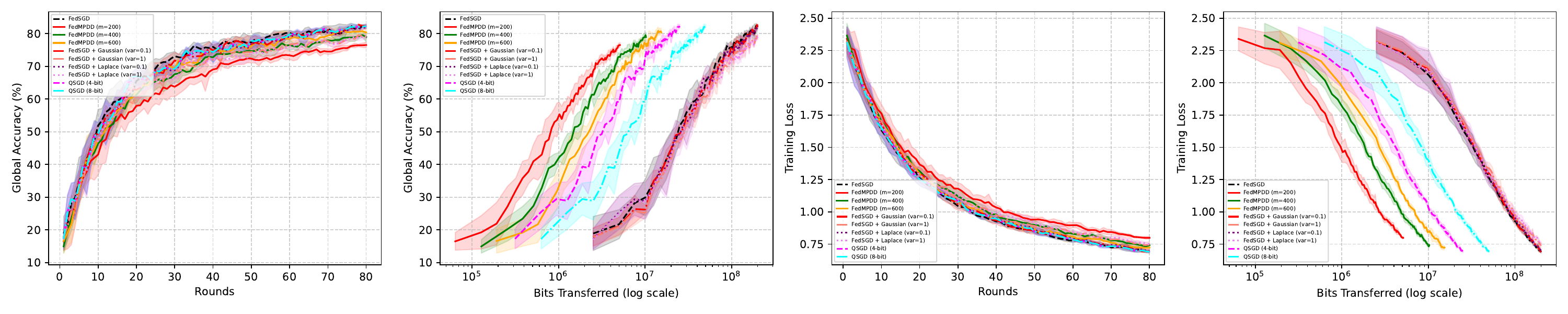}
\caption{{\small Training loss and accuracy curves versus communication rounds and number of transmitted bits for the logistic model on the MNIST dataset (i.i.d.).}}
\label{fig:logistic_mnist}
\end{figure*}

\begin{figure*}[t]
\centering
    \includegraphics[scale=0.3]{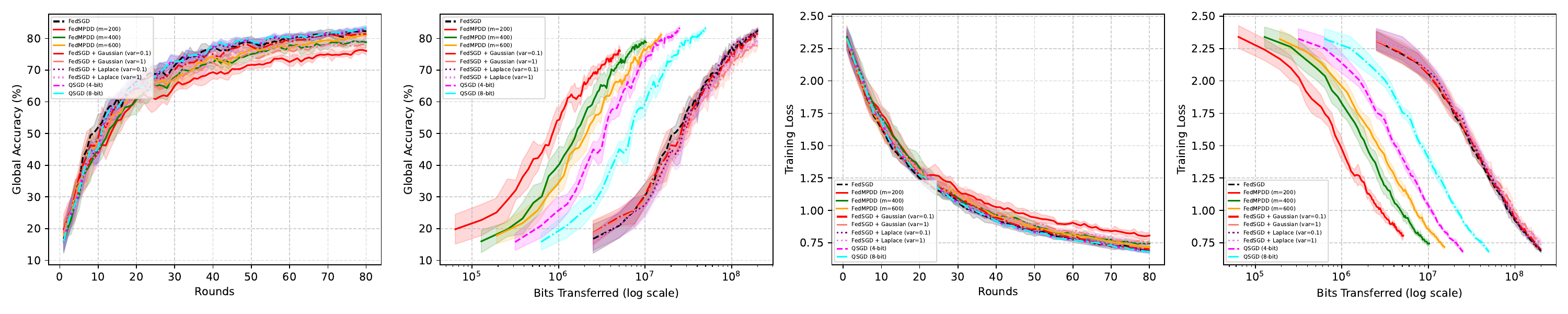}
    \caption{{\small Training loss and accuracy curves versus communication rounds and number of transmitted bits for the logistic model on the MNIST dataset (non-i.i.d.).}}
    \label{fig:non-idd-mnist}
\end{figure*}

\begin{figure*}[t]
\centering
    \includegraphics[scale=0.3]{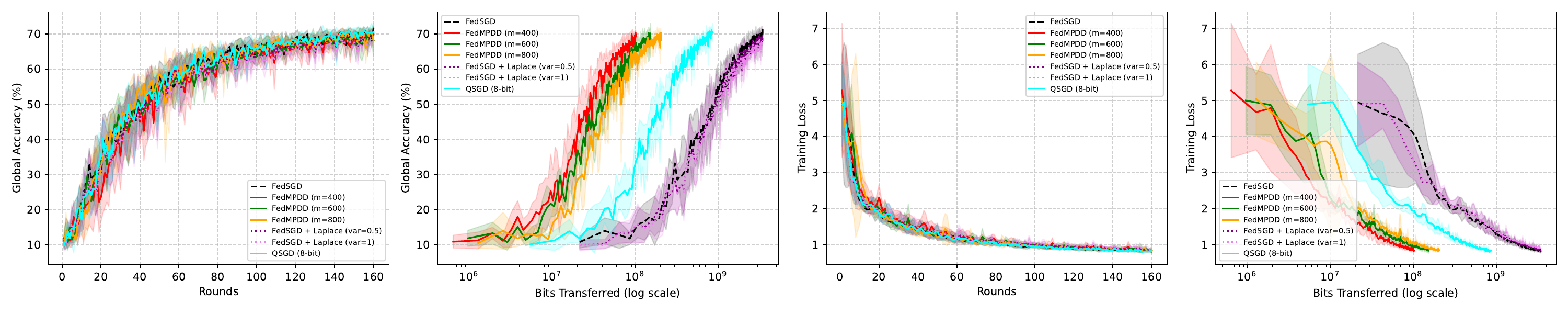}
    \caption{{\small Training loss and accuracy curves versus communication rounds and number of transmitted bits for the LeNet model on the FMNIST dataset (i.i.d.).}.}
    \label{fig:iid:lenet}
\end{figure*}

\begin{figure*}[t]
\centering
    \includegraphics[scale=0.3]{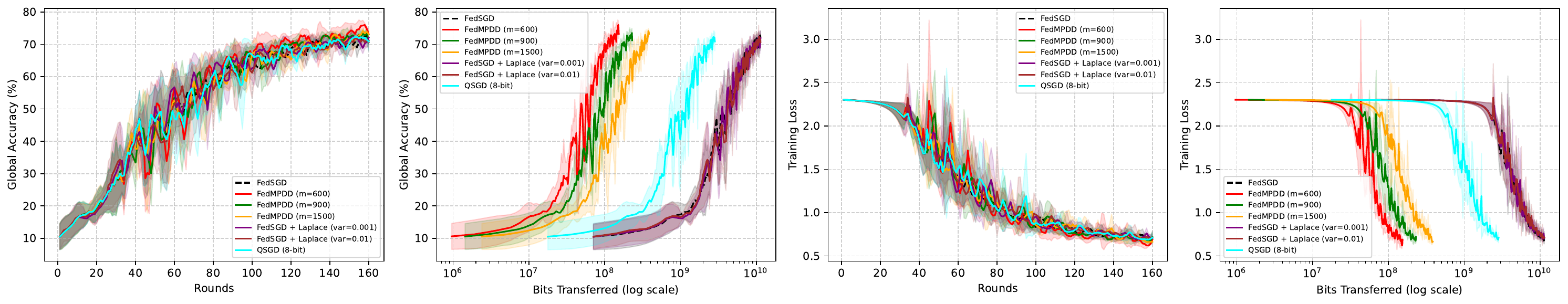}
    \caption{{\small Training loss and accuracy curves versus communication rounds and number of transmitted bits for the CNN model from \cite{lin2022personalized} on the FMNIST dataset (i.i.d.)}.}
    \label{fig:iid:fmnist}
\end{figure*}

\begin{figure*}[t]
\centering
    \includegraphics[scale=0.3]{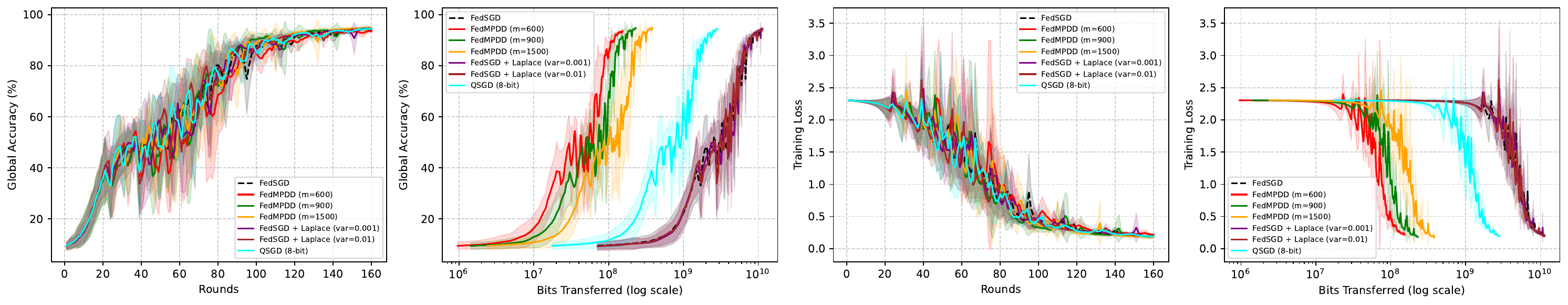}
    \caption{{\small Training loss and accuracy curves versus communication rounds and number of transmitted bits for the CNN model from \cite{lin2022personalized} on the MNIST dataset (i.i.d.)}.}
    \label{fig:cnn:mnist}
\end{figure*}

\begin{figure*}[t]
\centering
    \includegraphics[scale=0.3]{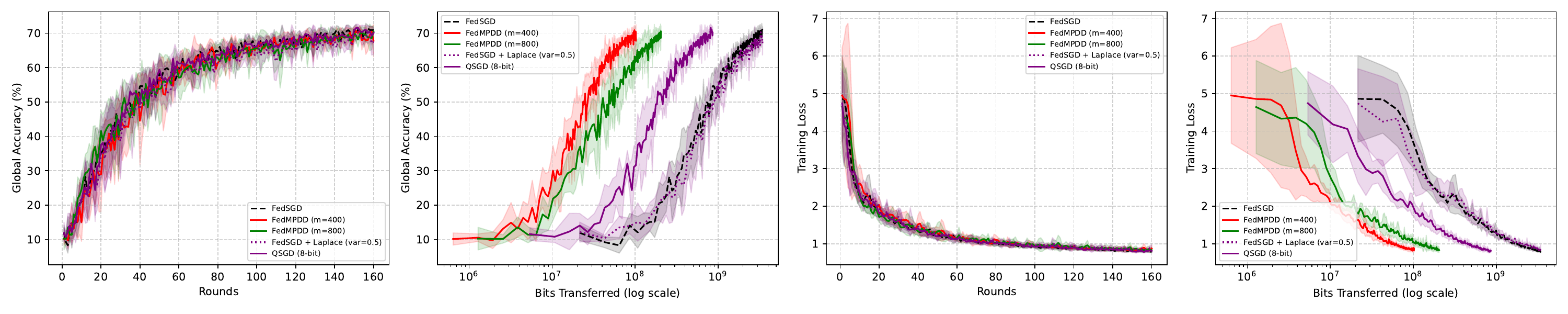}
    \caption{{\small Training loss and accuracy curves versus communication rounds and number of transmitted bits for the LeNet model on the FMNIST dataset (non-i.i.d.)}.}
    \label{fig:non-idd:lenet}
\end{figure*}

Figures \ref{fig:logistic_mnist}-\ref{fig:cnn_fmnist_non_iid} provide a comprehensive comparison between the proposed \texttt{FedMPDD}
algorithm and standard communication-efficient baselines across multiple datasets,
models, and data heterogeneity settings. In particular, these figures report
(i) test accuracy versus communication rounds, (ii) test accuracy versus total
transmitted bits, (iii) training loss versus communication rounds, and
(iv) training loss versus transmitted bits.

Figures~\ref{fig:logistic_mnist} and~\ref{fig:non-idd-mnist} focus on the logistic regression model trained on MNIST under i.i.d. and
non-i.i.d. data distributions, respectively. These results highlight that \texttt{FedMPDD}
achieves comparable or higher accuracy than competing methods while requiring
substantially fewer communicated bits. Moreover, the training loss curves demonstrate faster convergence of \texttt{FedMPDD} in terms of
both rounds and communication cost.

Figures~\ref{fig:iid:lenet} and~\ref{fig:non-idd:lenet} present results for the LeNet model on FMNIST under i.i.d. and non-i.i.d.
settings. Across all metrics, \texttt{FedMPDD} consistently improves the
accuracy--communication trade-off and exhibits smoother and more stable training loss
decay compared to gradient compression and noise-based baselines.

Figures~\ref{fig:iid:fmnist},~\ref{fig:cnn:mnist},~\ref{fig::mnist:Cnn}, and~\ref{fig:cnn_fmnist_non_iid} report analogous experiments for CNN models on the FMNIST and MNIST datasets. In these higher-dimensional settings, the benefits of \texttt{FedMPDD} become even more pronounced: the proposed method achieves similar final accuracy with significantly lower communication budgets and reaches low training loss values earlier in terms of transmitted bits. This behavior confirms that directional-derivative-based compression is particularly effective for large models.

Overall, the results in Figures~4–12 consistently demonstrate that \texttt{FedMPDD}
offers a favorable trade-off between accuracy, convergence speed, and communication
efficiency across different models, datasets, and levels of data heterogeneity.

Table~\ref{tab:fedmpdd_all_methods:mnist_idd}, Figure~\ref{fig:mnist:recovery}, and Figure~\ref{fig:ssim_mnist} jointly evaluate the privacy and reconstruction
behavior of the proposed \texttt{FedMPDD} algorithm compared to standard
communication-efficient baselines under gradient inversion attacks.
Table~VI reports quantitative results in terms of test accuracy under a fixed
communication budget, communication usage to reach a target accuracy, defendability,
and reconstruction quality measured by SSIM.

As shown in Table~\ref{tab:fedmpdd_all_methods:mnist_idd}, classical baselines such as FedSGD, FedSGD with Gaussian or
Laplace noise, and QSGD either exceed the communication budget early or remain
vulnerable to gradient inversion attacks, as indicated by high SSIM values and
unsuccessful defendability. In contrast, \texttt{FedMPDD} achieves competitive test
accuracy while using significantly fewer communicated bytes and consistently
maintaining successful defense against reconstruction attacks.

These observations are further supported by the qualitative reconstruction results in
Figure~\ref{fig:mnist:recovery}. While the attacker is able to recover visually accurate digit samples for
FedSGD, noise-perturbed FedSGD, and QSGD, the reconstructed images corresponding to
\texttt{FedMPDD} appear highly noisy and uninformative, even for larger projection
dimensions. This visual degradation indicates that the transmitted directional
derivatives do not reveal sufficient information for successful data recovery.

Figure~\ref{fig:ssim_mnist} quantitatively summarizes the reconstruction quality using SSIM.
Baseline methods exhibit high SSIM values close to one, confirming near-perfect
reconstruction. In contrast, \texttt{FedMPDD} yields significantly lower SSIM values
across all projection dimensions, demonstrating strong resistance to gradient
inversion. Together, these results confirm that \texttt{FedMPDD} provides a favorable
trade-off between accuracy, communication efficiency, and privacy preservation.

Table~\ref{tab:fedmpdd_all_methods_table_fmnist_iid},
Figure~\ref{fig:recovery_fmnist}, and
Figure~\ref{fig:ssim:fmnist}
present a comprehensive evaluation of privacy leakage and reconstruction robustness
for the LeNet model trained on the FMNIST dataset under the i.i.d. setting.
The comparison includes test accuracy under a fixed communication budget,
communication usage to reach a target accuracy, defendability against gradient
inversion attacks, and reconstruction quality measured by SSIM.

As reported in Table~\ref{tab:fedmpdd_all_methods_table_fmnist_iid}, baseline methods
such as FedSGD and QSGD either fail to provide sufficient privacy protection or incur
substantially higher communication costs. While adding Laplace noise to FedSGD reduces
reconstruction quality, this comes at the expense of degraded accuracy and increased
communication usage. In contrast, \texttt{FedMPDD} achieves significantly higher test
accuracy while requiring markedly fewer transmitted bytes and consistently satisfying
the defendability criterion across all tested projection dimensions.

The qualitative reconstruction results in
Figure~\ref{fig:recovery_fmnist} further illustrate these findings.
For FedSGD and QSGD, the attacker is able to recover visually meaningful input images,
indicating severe privacy leakage. In comparison, the reconstructed images
corresponding to \texttt{FedMPDD} appear highly noisy and unstructured, even for larger
values of the projection dimension, demonstrating that the shared directional
derivatives do not reveal sufficient information for successful data recovery.

This behavior is quantitatively confirmed by the SSIM results in
Figure~\ref{fig:ssim:fmnist}. While FedSGD and QSGD exhibit SSIM values close to one,
indicating near-perfect reconstruction, \texttt{FedMPDD} consistently yields SSIM
values close to zero. These results confirm that \texttt{FedMPDD} provides strong
resistance to gradient inversion attacks while maintaining favorable
accuracy–communication trade-offs.

Tables~\ref{tab:fedmpdd_lenet_mnist_noniid},
\ref{tab:fedmpdd_lenet_fmnist_noniid},
\ref{tab:fedmpdd_cnn_fmnist_iid},
\ref{tab:fedmpdd_cnn_mnist_noniid}, and
\ref{tab:fedmpdd_cnn_fmnist_noniid}
summarize the accuracy, communication cost, and privacy behavior of different
methods across non-i.i.d. data distributions and deeper CNN architectures.
All results are reported under a fixed communication budget, together with the
communication usage required to reach a target accuracy and the defendability
against gradient inversion attacks.

Across all settings, standard baselines such as FedSGD and QSGD either fail to achieve
competitive test accuracy within the communication budget or remain vulnerable to
privacy leakage, as indicated by unsuccessful defendability.
Adding Laplace noise to FedSGD improves privacy in some cases, but this typically
comes at the expense of degraded accuracy and substantially increased communication
usage.

In contrast, \texttt{FedMPDD} consistently achieves significantly higher test accuracy
while using orders of magnitude fewer communicated bytes.
This behavior is observed for both LeNet and CNN models, and persists under
non-i.i.d. data distributions on MNIST and FMNIST.
Moreover, \texttt{FedMPDD} satisfies the defendability criterion in all reported
settings, demonstrating robustness to gradient inversion attacks without relying on
explicit noise injection.

Notably, as the projection dimension increases, \texttt{FedMPDD} exhibits a smooth
accuracy--communication trade-off: larger projection dimensions improve accuracy at
the cost of increased communication, while still maintaining strong privacy
guarantees.
These results confirm that \texttt{FedMPDD} scales favorably to deeper models and
heterogeneous data while simultaneously addressing communication efficiency and
privacy preservation.

Table~\ref{tab:fedmpdd_all_methods_cnn_mnist_iid},
Figure~\ref{fig:revocery:cnn:mnist}, and
Figure~\ref{fig:ssim:cnn:mnist}
evaluate the privacy leakage and reconstruction robustness of different
communication-efficient methods for the CNN model on the MNIST dataset under the i.i.d.
setting. The comparison includes test accuracy under a fixed communication budget,
communication usage to reach a target accuracy, defendability against gradient
inversion attacks, and reconstruction quality measured by SSIM.

As shown in Table~\ref{tab:fedmpdd_all_methods_cnn_mnist_iid}, baseline methods such as
FedSGD and QSGD either exhibit poor test accuracy under the fixed communication budget
or remain vulnerable to gradient inversion attacks, as reflected by high SSIM values
and unsuccessful defendability. Adding Laplace noise to FedSGD reduces reconstruction
quality but requires substantially higher communication usage and still fails to
achieve a favorable accuracy–communication trade-off. In contrast, \texttt{FedMPDD}
achieves significantly higher test accuracy while using orders of magnitude fewer
transmitted bytes and consistently satisfying the defendability criterion across all
tested projection dimensions.

The qualitative reconstruction results in
Figure~\ref{fig:revocery:cnn:mnist} further support these findings.
For FedSGD, noise-perturbed FedSGD, and QSGD, the attacker is able to recover visually
meaningful digit structures, indicating severe privacy leakage. In comparison, the
reconstructed images corresponding to \texttt{FedMPDD} are highly distorted and lack
semantic structure, demonstrating that the compressed directional information is
insufficient for successful input recovery.

Finally, Figure~\ref{fig:ssim:cnn:mnist} reports SSIM as a function of attack iteration.
Baseline methods rapidly converge to high SSIM values, indicating increasingly accurate
reconstructions as the attack progresses. In contrast, \texttt{FedMPDD} consistently
maintains low SSIM values across iterations, confirming sustained resistance to
gradient inversion attacks throughout the optimization process. Together, these
results demonstrate that \texttt{FedMPDD} provides strong privacy protection while
preserving favorable accuracy and communication efficiency for CNN models.

Figure~\ref{fig:k=600} illustrates the evolution of reconstruction quality over
training epochs for the LeNet model under a gradient inversion attack.
Specifically, the figure reports the SSIM values produced by the attack in~\cite{yu2025gi}
when applied to the projected directional derivative estimator used in
\texttt{FedMPDD} with projection dimension $m=600$.
Across all training epochs, the SSIM values remain consistently low and do not
exhibit any increasing trend, indicating that the attacker is unable to recover
meaningful information about the private inputs as training progresses.
This result demonstrates that \texttt{FedMPDD} enforces a stable and persistent
privacy guarantee throughout the entire training process, rather than relying on
privacy that degrades over time.

Table~\ref{tab:fedmpdd_m_sweep} illustrates the impact of the number of random
directions $m$ on the test accuracy of \texttt{FedMPDD} for the LeNet model on MNIST.
As predicted in Remark~\ref{remak:choosing_m}, increasing $m$ improves accuracy by reducing the distortion
introduced by JL Lemma.
Specifically, smaller values of $m$ lead to higher projection error and consequently
degraded optimization performance, while larger values of $m$ yield progressively
better accuracy as the projected directional derivative more faithfully preserves
gradient information.

Importantly, the observed accuracy gains saturate as $m$ increases, indicating a
clear communication--accuracy trade-off.
This behavior is consistent with the JL embedding guarantee, where choosing $m$ within
the theoretically valid regime is sufficient to control the distortion parameter
$\varepsilon$ and ensure stable convergence.
These results confirm that selecting $m$ according to the JL scaling in Remark~\ref{remak:choosing_m} yields
both strong empirical performance and adherence to the theoretical guarantees.

Table~\ref{tab:mpdd_latency} reports the average per-round, per-client latency required
to compute the directional derivative encodings in \texttt{FedMPDD} for different
numbers of random directions $m$ on the LeNet--MNIST setup.
As shown, even for larger values of $m$, the measured client-side computation time
remains on the order of milliseconds, confirming that the encoding step does not
constitute a computational bottleneck in practice. These empirical results are consistent with Remark~\ref{remark:broader_impact}.
Although the theoretical encoding cost of \texttt{FedMPDD} scales as $\mathcal{O}(dm)$, this cost
is negligible relative to the overall training pipeline in our experiments.
Moreover, as discussed in Remark~\ref{remark:broader_impact}, the directional
derivative computation can be efficiently implemented using Jacobian--vector
products, avoiding explicit full-gradient computation. This projected-forward strategy further reduces practical overhead and can even lead to computational savings in deep models, making \texttt{FedMPDD} well suited for
resource-constrained client devices.

\twocolumn

\begin{figure*}[t]
\centering
    \includegraphics[scale=0.3]{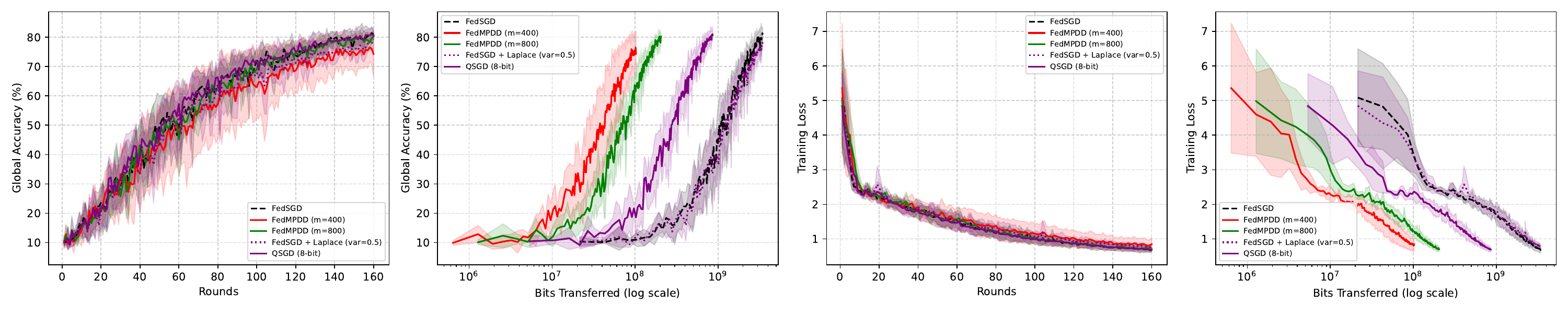}
    \caption{{\small Training loss and accuracy curves versus communication rounds and number of transmitted bits for the LeNet model on the MNIST dataset (non-i.i.d.)}.}
    \label{fig}
\end{figure*}

\begin{figure*}[t]
\centering
    \includegraphics[scale=0.3]{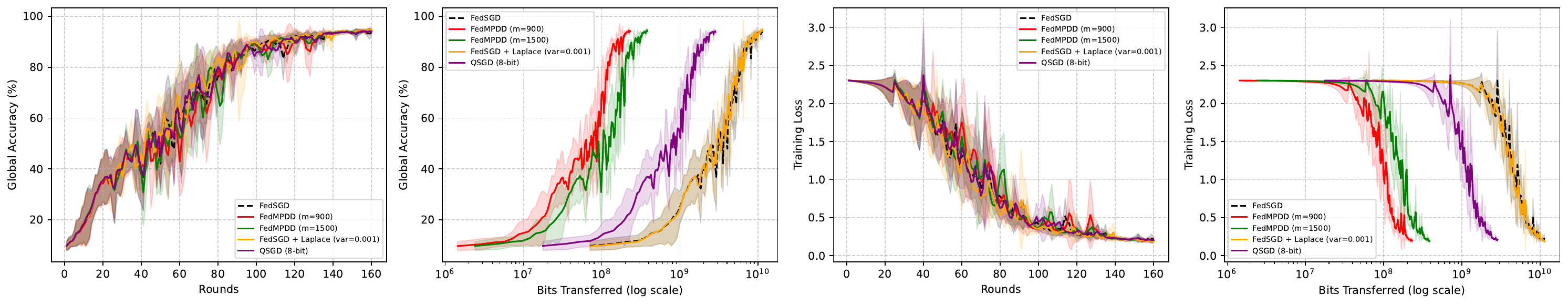}
    \caption{{\small Training loss and accuracy curves versus communication rounds and number of transmitted bits for the CNN model from \cite{lin2022personalized} on the MNIST dataset (non-i.i.d.)}.}
    \label{fig::mnist:Cnn}
\end{figure*}

\begin{figure*}[t]
\centering
    \includegraphics[scale=0.3]{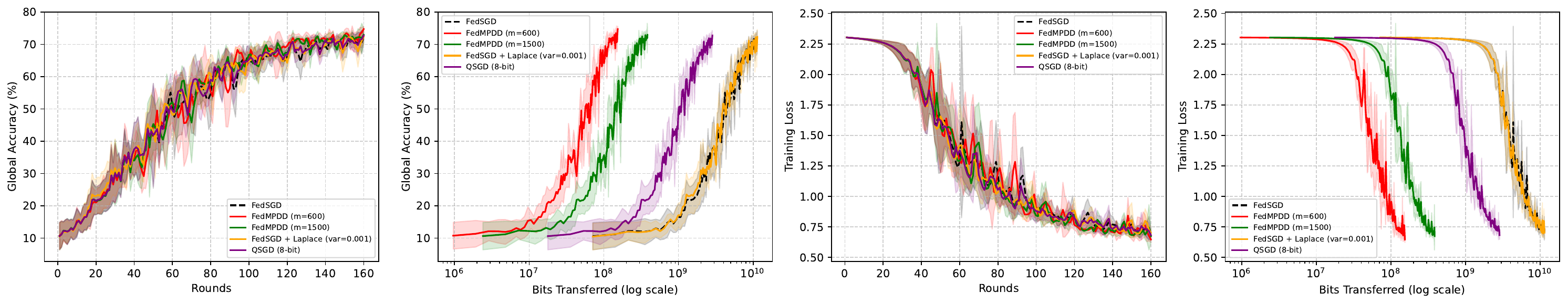}
    \caption{{\small Training loss and accuracy curves versus communication rounds and number of transmitted bits for the CNN model from \cite{lin2022personalized} on the FMNIST dataset (non-i.i.d.)}.}
    \label{fig:cnn_fmnist_non_iid}
\end{figure*}

\onecolumn

\twocolumn

\begin{table*}[t]
\centering
\footnotesize
\caption{Comparison of test accuracy under a fixed communication budget, communication usage for target accuracy, privacy leakage, and reconstruction quality on MNIST (i.i.d.) using the attack of \cite{zhu2019deep} with a logistic model. A $\star$ in the Test Acc column indicates that the communication budget was already exceeded in the first iteration of the algorithm.}
\label{tab:fedmpdd_all_methods:mnist_idd}
\resizebox{\linewidth}{!}{%
\begin{tabular}{lcc|cc|cc}
\toprule
\textbf{Method} & \textbf{Bytes Budget} & \textbf{Test Acc} & \textbf{Target Acc} & \textbf{Used Bytes} & \textbf{Defendability} & \textbf{SSIM} \\
\midrule
FedSGD                              & 2000000 & $\star$          & 60 & 40192000.0 & \xmark{} & 1.00 \\
FedSGD + Gaussian (var=0.1)         & 2000000 & $\star$           & 60 & 42704000.0 & \xmark{} & 0.80 \\
FedSGD + Gaussian (var=1)           & 2000000 & $\star$           & 60 & 45216000.0 & \xmark{} & 0.59 \\
FedSGD + Laplace (var=0.1)          & 2000000 & $\star$           & 60 & 42704000.0 & \xmark{} & 0.82 \\
FedSGD + Laplace (var=1)            & 2000000 & $\star$           & 60 & 45216000.0 & \xmark{} & 0.60 \\
\textbf{FedMPDD (m=200)}            & 2000000 & \textbf{65.29} & 60 & \textbf{1536000.0} & \cmark{} & 0.03 \\
\textbf{FedMPDD (m=400)}            & 2000000 & \textbf{57.62} & 60 & \textbf{2432000.0} & \cmark{} & 0.14 \\
\textbf{FedMPDD (m=600)}            & 2000000 & \textbf{48.85} & 60 & \textbf{3456000.0} & \cmark{} & 0.13 \\
QSGD (4-bit)                        & 2000000 & 38.92          & 60 & 5343440.0 & \xmark{} & 0.88 \\
QSGD (8-bit)                        & 2000000 & 28.86          & 60 & 10681440.0 & \xmark{} & 0.99 \\
\bottomrule
\end{tabular}%
}
\vspace{0.5em}
\end{table*}

\begin{figure*}[t]
\centering
    \includegraphics[scale=0.2]{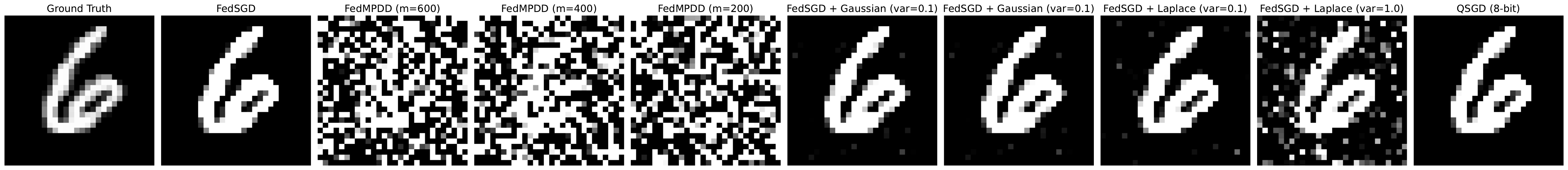}
    \caption{{\small Attack results on logistic model using MNIST dataset}.}
    \label{fig:mnist:recovery}
\end{figure*}

\begin{figure*}[t]
\centering
    \includegraphics[scale=0.45]{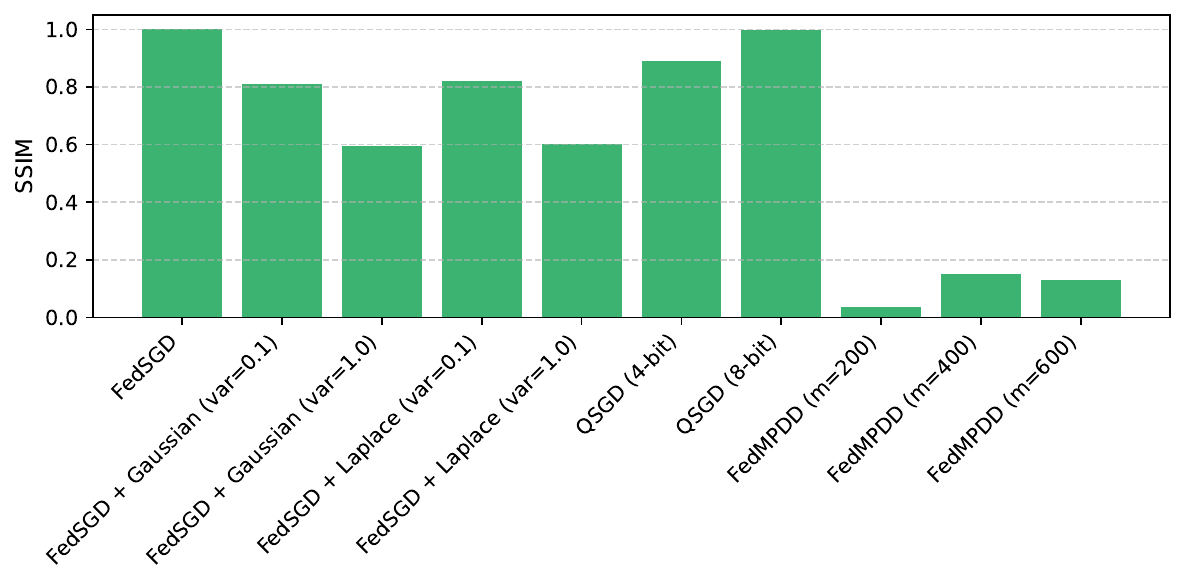}
    \caption{{\small SSIM bar plot on logistic model using MNIST dataset}.}
    \label{fig:ssim_mnist}
\end{figure*}

\onecolumn

\twocolumn

\begin{table*}[t]
\centering
\footnotesize
\caption{Comparison of test accuracy under a fixed communication budget, communication usage for target accuracy, privacy leakage, and reconstruction quality on FMNIST (i.i.d.) using the attack of \cite{zhu2019deep} with LeNet model. }
\label{tab:fedmpdd_all_methods_table_fmnist_iid}
\resizebox{\linewidth}{!}{%
\begin{tabular}{lcc|cc|cc}
\toprule
\textbf{Method} & \textbf{Bytes Budget} & \textbf{Test Acc} & \textbf{Target Acc} & \textbf{Used Bytes} & \textbf{Defendability} & \textbf{SSIM} \\
\midrule
FedSGD                              & 90000000 & 15.85          & 60 & 1331859200.0 & \xmark{} & 1.00 \\
FedSGD + Laplace (var=0.5)          & 90000000 & 15.76          & 60 & 1245932800.0 & \cmark{} & $\ll 0.03$ \\
FedSGD + Laplace (var=1)            & 90000000 & 15.91          & 60 & 1525193600.0 & \cmark{} & $\ll 0.03$ \\
\textbf{FedMPDD (m=400)}            & 90000000 & \textbf{66.77} & 60 & \textbf{44160000.0} & \cmark & $\ll 0.03$ \\
\textbf{FedMPDD (m=600)}            & 90000000 & \textbf{66.10} & 60 & \textbf{61440000.0} & \cmark & $\ll 0.03$ \\
\textbf{FedMPDD (m=800)}            & 90000000 & \textbf{64.84} & 60 & \textbf{78080000.0} & \cmark & $\ll 0.03$ \\
QSGD (8-bit)                        & 90000000 & 25.73          & 60 & 311576000.0 & \xmark{} & 0.99 \\
\bottomrule
\end{tabular}%
}
\vspace{0.5em}
\end{table*}

\begin{figure*}[t]
\centering
    \includegraphics[scale=0.25]{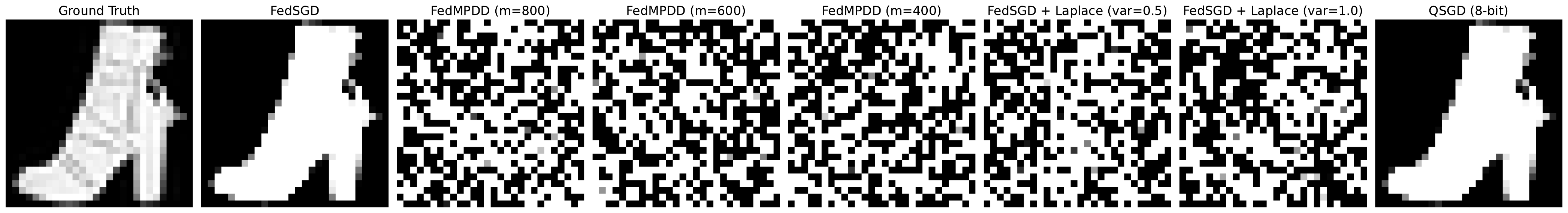}
    \caption{{\small Attack results on LeNet using FMNIST dataset}.}
    \label{fig:recovery_fmnist}
\end{figure*}

\begin{figure*}[t]
\centering
    \includegraphics[scale=0.4]{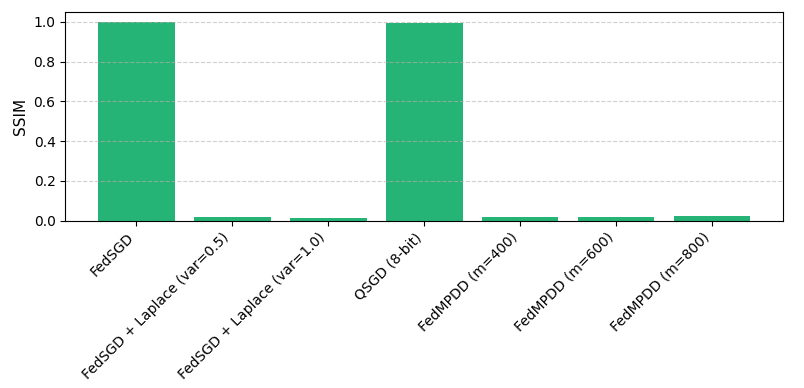}
    \caption{{\small SSIM bar plot on LeNet using FMNIST dataset}.}
    \label{fig:ssim:fmnist}
\end{figure*}


\clearpage

\clearpage


\twocolumn

\begin{table*}[t]
\centering
\footnotesize
\caption{Comparison of test accuracy under a fixed communication budget, communication usage for target accuracy, privacy leakage, and reconstruction quality on MNIST (i.i.d.) using the attack of \cite{yu2025gi} with CNN model from \cite{lin2022personalized}. }
\label{tab:fedmpdd_all_methods_cnn_mnist_iid}
\resizebox{\linewidth}{!}{%
\begin{tabular}{lcc|cc|cc}
\toprule
\textbf{Method} & \textbf{Bytes Budget} & \textbf{Test Acc} & \textbf{Target Acc} & \textbf{Used Bytes} & \textbf{Defendability} & \textbf{SSIM} \\
\midrule
FedSGD                              & 150000000 & 10.40          & 60 & 3980569600.0 & \xmark & 0.81 \\
FedSGD + Laplace (var=0.001)          & 150000000 & 10.30          & 60 & 4193814400.0 & \xmark & 0.48 \\
FedSGD + Laplace (var=0.01)            & 150000000 & 10.33          & 60 & 3625161600.0 & \cmark & 0.13 \\
\textbf{FedMPDD (m=600)}            & 150000000 & \textbf{93.26} & 60 & \textbf{64320000.0} & \cmark & 0.17 \\
\textbf{FedMPDD (m=900)}            & 150000000 & \textbf{91.46} & 60 & \textbf{86400000.0} & \cmark & 0.28 \\
\textbf{FedMPDD (m=1500)}            & 150000000 & \textbf{59.46} & 60 & \textbf{146400000.0} & \cmark & 0.28 \\
QSGD (8-bit)                        & 150000000 & 14.56          & 60 & 977460000.0 & \xmark & 0.64 \\
\bottomrule
\end{tabular}%
}
\vspace{0.5em}
\end{table*}

\begin{figure*}[t]
\centering
    \includegraphics[scale=0.25]{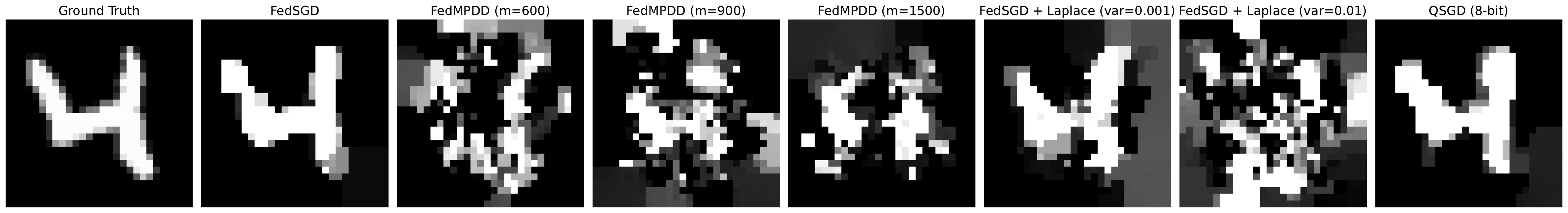}
    \caption{{\small Attack results on CNN model from \cite{lin2022personalized} using MNIST dataset}.}
    \label{fig:revocery:cnn:mnist}
\end{figure*}

\begin{figure*}[t]
\centering
    \includegraphics[scale=0.4]{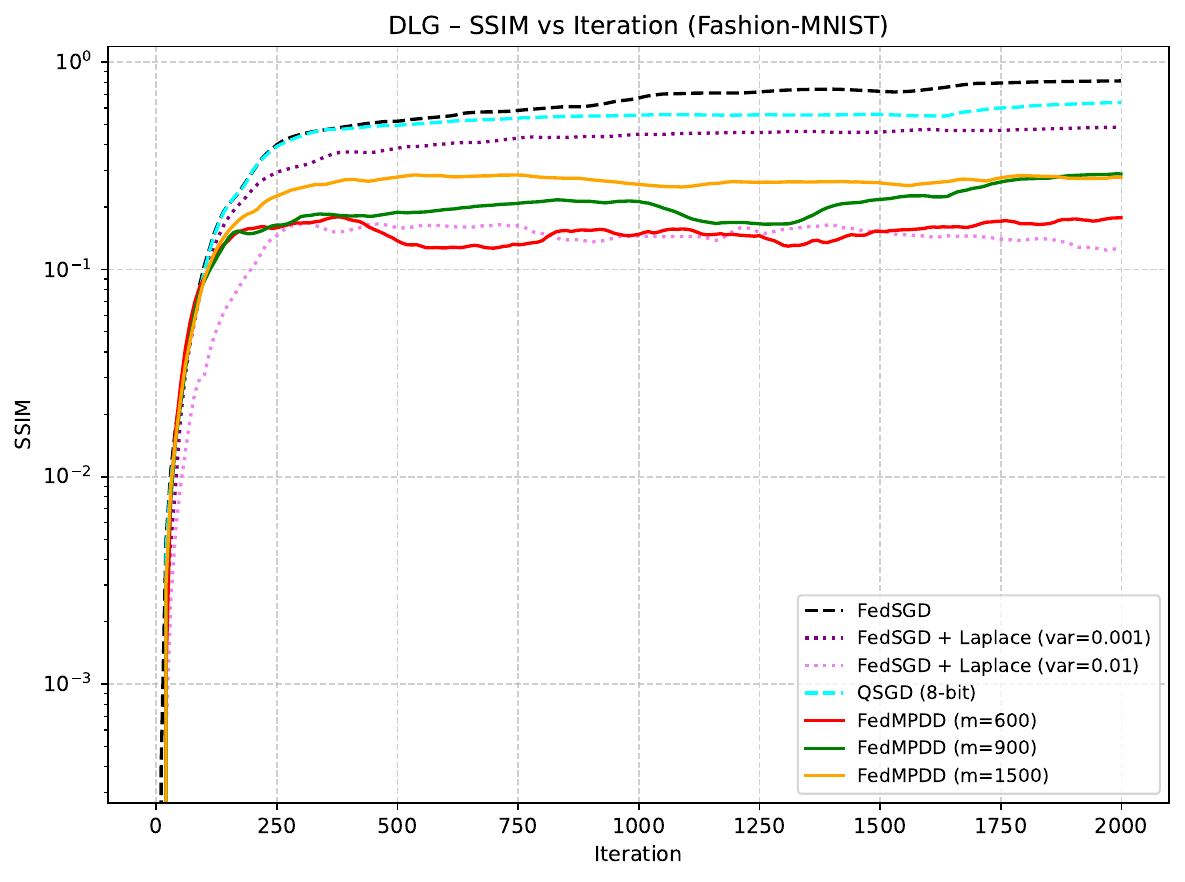}
    \caption{{\small SSIM plot versus iteration on CNN model from \cite{lin2022personalized} using MNIST dataset}.}
    \label{fig:ssim:cnn:mnist}
\end{figure*}



\clearpage

\clearpage
\onecolumn

\twocolumn

\begin{table*}[t]
\centering
\footnotesize
\caption{Comparison of test accuracy under a fixed communication budget, communication usage for target accuracy, and privacy leakage on MNIST (non-i.i.d.) using the attack of \cite{yu2025gi} with LeNet model.}
\label{tab:fedmpdd_lenet_mnist_noniid}
\resizebox{\linewidth}{!}{%
\begin{tabular}{lcc|cc|c}
\toprule
\textbf{Method} & \textbf{Bytes Budget} & \textbf{Test Acc} & \textbf{Target Acc} & \textbf{Used Bytes} & \textbf{Defendability} \\
\midrule
FedSGD                              & 90000000 & 11.06          & 60 & 1460748800.0 & \xmark \\
FedSGD + Laplace (var=0.5)          & 90000000 & 12.75          & 60 & 1482230400.0 & \cmark \\
\textbf{FedMPDD (m=400)}            & 90000000 & \textbf{76.00} & 60 & \textbf{51840000.0} & \cmark\\
\textbf{FedMPDD (m=800)}            & 90000000 & \textbf{57.80} & 60 & \textbf{93440000.0} & \cmark \\
QSGD (8-bit)                        & 90000000 & 18.66          & 60 & 359924000.0 & \xmark \\
\bottomrule
\end{tabular}%
}
\vspace{0.5em}
\end{table*}


\begin{table*}[t]

\centering
\footnotesize
\caption{Comparison of test accuracy under a fixed communication budget, communication usage for target accuracy, and privacy leakage on FMNIST (non-i.i.d.) using the attack of \cite{zhu2019deep} with LeNet model.}
\label{tab:fedmpdd_lenet_fmnist_noniid}

\resizebox{\linewidth}{!}{%
\begin{tabular}{lcc|cc|c}
\toprule
\textbf{Method} & \textbf{Bytes Budget} & \textbf{Test Acc} & \textbf{Target Acc} & \textbf{Used Bytes} & \textbf{Defendability} \\
\midrule
FedSGD                              & 90000000 & 12.05          & 60 & 1288896000.0 & \xmark \\
FedSGD + Laplace (var=0.5)          & 90000000 & 15.06          & 60 & 1224451200.0 & \cmark \\
\textbf{FedMPDD (m=400)}            & 90000000 & \textbf{69.47} & 60 & \textbf{37760000.0} & \cmark \\
\textbf{FedMPDD (m=800)}            & 90000000 & \textbf{61.18} & 60 & \textbf{74240000.0} & \cmark \\
QSGD (8-bit)                        & 90000000 & 24.30          & 60 & 306204000.0 & \xmark \\
\bottomrule
\end{tabular}%
}
\vspace{0.5em}
\end{table*}

\clearpage


\clearpage

\clearpage

\begin{table*}[t]

\centering
\footnotesize
\caption{Comparison of test accuracy under a fixed communication budget, communication usage for target accuracy, privacy leakage, and reconstruction quality on FMNIST (i.i.d.) using the attack of \cite{yu2025gi} with CNN model from \cite{lin2022personalized}.}
\label{tab:fedmpdd_cnn_fmnist_iid}
\resizebox{\linewidth}{!}{%
\begin{tabular}{lcc|cc|c}
\toprule
\textbf{Method} & \textbf{Bytes Budget} & \textbf{Test Acc} & \textbf{Target Acc} & \textbf{Used Bytes} & \textbf{Defendability} \\
\midrule
FedSGD                         & 150000000   & 11.74          & 60 & 5757609600.0   & \xmark \\
FedSGD + Laplace (var=0.001)   & 150000000   & 11.95          & 60 & 4975712000.0   & \cmark \\
FedSGD + Laplace (var=0.01)    & 150000000   & 12.48          & 60 & 5402201600.0   & \cmark \\
\textbf{FedMPDD (m=600)}       & 150000000   & \textbf{75.81} & 60 & \textbf{73920000.0}  & \cmark \\
\textbf{FedMPDD (m=900)}       & 150000000   & \textbf{62.36} & 60 & \textbf{103680000.0} & \cmark \\
\textbf{FedMPDD (m=1500)}      & 150000000   & \textbf{47.32} & 60 & \textbf{199200000.0} & \cmark \\
QSGD (8-bit)                   & 150000000   & 15.51          & 60 & 1421760000.0   & \xmark \\
\bottomrule
\end{tabular}%
}
\vspace{0.5em}
\end{table*}

\begin{table*}[t]

\centering
\footnotesize
\caption{Comparison of test accuracy under a fixed communication budget, communication usage for target accuracy, and privacy leakage on MNIST (non-i.i.d.) using the attack of \cite{zhu2019deep} with CNN model from \cite{lin2022personalized}.}
\label{tab:fedmpdd_cnn_mnist_noniid}
\resizebox{\linewidth}{!}{%
\begin{tabular}{lcc|cc|c}
\toprule
\textbf{Method} & \textbf{Bytes Budget} & \textbf{Test Acc} & \textbf{Target Acc} & \textbf{Used Bytes} & \textbf{Defendability} \\
\midrule
FedSGD                         & 200000000   & 11.40          & 60 & 3696243200.0   & \xmark \\
FedSGD + Laplace (var=0.001)   & 200000000   & 11.04          & 60 & 3696243200.0   & \xmark \\
\textbf{FedMPDD (m=900)}       & 200000000   & \textbf{93.08} & 60 & \textbf{90720000.0}  & \cmark \\
\textbf{FedMPDD (m=1500)}      & 200000000   & \textbf{80.67} & 60 & \textbf{151200000.0} & \cmark \\
QSGD (8-bit)                   & 200000000   & 21.84          & 60 & 1030776000.0   & \xmark \\
\bottomrule
\end{tabular}%
}
\vspace{0.5em}
\end{table*}

\begin{table*}[t]

\centering
\footnotesize
\caption{Comparison of test accuracy under a fixed communication budget, communication usage for target accuracy, privacy leakage, and reconstruction quality on FMNIST (non-i.i.d.) using the attack of \cite{zhu2019deep} with CNN model from \cite{lin2022personalized}.}
\label{tab:fedmpdd_cnn_fmnist_noniid}
\resizebox{\linewidth}{!}{%
\begin{tabular}{lcc|cc|c}
\toprule
\textbf{Method} & \textbf{Bytes Budget} & \textbf{Test Acc} & \textbf{Target Acc} & \textbf{Used Bytes} & \textbf{Defendability} \\
\midrule
FedSGD                         & 150000000   & 12.22          & 60 & 5828691200.0   & \xmark \\
FedSGD + Laplace (var=0.001)   & 150000000   & 12.03          & 60 & 5615446400.0   & \cmark \\
\textbf{FedMPDD (m=600)}       & 150000000   & \textbf{70.83} & 60 & \textbf{74880000.0}  & \cmark \\
\textbf{FedMPDD (m=1500)}      & 150000000   & \textbf{51.32} & 60 & \textbf{199200000.0} & \cmark \\
QSGD (8-bit)                   & 150000000   & 13.51          & 60 & 1457304000.0   & \xmark \\
\bottomrule
\end{tabular}%
}
\vspace{0.5em}
\end{table*}

\clearpage

\begin{figure}[t]
\centering
    \includegraphics[scale=0.35]{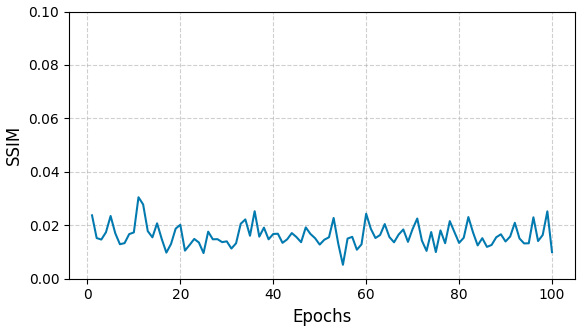}
    \caption{\small{The SSIM scores produced by the attack in \cite{yu2025gi} on the LeNet model using the \emph{projected directional derivative} estimator \(\hat{\vect{g}}_i(\vect{x})\) with \(m=600\) in \texttt{FedMPDD} remain uniformly low over 100 training epochs, confirming that \texttt{FedMPDD} enforces a constant privacy guarantee throughout the entire training process.
}}
    \label{fig:k=600}
\end{figure}

\vspace{0.1in}

\begin{table}[t]
\centering
\scriptsize
\caption{\small{\texttt{FedMPDD} performance with varying numbers of random directions $m$ on the MNIST dataset using the LeNet model.}}
\label{tab:fedmpdd_m_sweep}
\begin{tabular}{@{}l S[table-format=2.2]@{}}
\toprule
\textbf{Method} & {\textbf{Test Acc. (\%)}} \\
\midrule
\textbf{FedMPDD} (m=50)  & 30.44 \\
\textbf{FedMPDD} (m=200) & 75.02 \\
\textbf{FedMPDD} (m=400) & 77.52 \\
\textbf{FedMPDD} (m=600) & 79.02 \\
\bottomrule
\end{tabular}
\end{table}

\vspace{0.1in}

\begin{table}[t]
\centering
\scriptsize
\caption{\small{Per-round, per-client latency (ms) to compute $\vect{s}_{i}^{k}[j] = (\vect{u}_{k,i}^{(j)})^\top \vect{g}_{i}(\vect{x}_{k})$,
averaged over clients and rounds on LeNet–MNIST.}}
\label{tab:mpdd_latency}
\begin{tabular}{@{}S[table-format=4.0] S[table-format=1.2]@{}}
\toprule
{\# random directions $m$} & {\textbf{Avg.\ latency (ms)}} \\
\midrule
400 & 0.43 \\
600 & 0.61 \\
800 & 0.93 \\
\bottomrule
\end{tabular}
\end{table}

\end{document}